\documentclass[10pt,twocolumn,letterpaper]{article}

\usepackage{cvpr}              

\usepackage{mdframed}

\usepackage[skip=0pt]{caption}

\definecolor{cvprblue}{rgb}{0.21,0.49,0.74}
\usepackage[pagebackref,breaklinks,colorlinks,allcolors=cvprblue]{hyperref}
\usepackage{booktabs}       
\usepackage{multirow}       
\usepackage{multicol}       
\usepackage{graphicx}       
\usepackage{makecell}       
\usepackage{array}          
\usepackage{rotating}       
\usepackage[table]{xcolor}  
\usepackage{colortbl}       
\usepackage{amsmath}        
\usepackage{pifont}         
\usepackage{caption}        

\usepackage[capitalize]{cleveref}
\crefname{section}{Sec.}{Secs.}
\Crefname{section}{Section}{Sections}
\Crefname{table}{Table}{Tables}
\crefname{table}{Tab.}{Tabs.}

\usepackage[ruled,vlined]{algorithm2e}

\definecolor{my_blue}{HTML}{d3eaf2}
\definecolor{mydarkgreen}{RGB}{0,100,0} 

\def\paperID{} 
\def\confName{CVPR}
\def\confYear{2026}

\title{The Devil Is in Gradient Entanglement: Energy-Aware Gradient Coordinator for Robust Generalized Category Discovery}

\author{ 
Haiyang Zheng$^{1}$ \enspace 
Nan Pu$^{2*}$ \enspace
Yaqi Cai$^{1}$ \enspace
Teng Long$^{1}$ \enspace
Wenjing Li$^{2}$ \enspace
Nicu Sebe$^{1}$ \enspace
Zhun Zhong$^{2}$\thanks{Corresponding author.} \\
$^{1}$University of Trento, Italy \enspace
$^{2}$Hefei University of Technology, China \\
{\tt\small 
\{haiyang.zheng, yaqi.cai, teng.long, niculae.sebe\}@unitn.it}\\
{\tt\small \{n.pu, wenjingli, zhunzhong\}@hfut.edu.cn
}
}

\begin{document}
\maketitle
\begin{abstract}
Generalized Category Discovery (GCD) leverages labeled data to categorize unlabeled samples from known or unknown classes. Most previous methods jointly optimize supervised and unsupervised objectives and achieve promising results. However, inherent optimization interference still limits their ability to improve further. 
Through quantitative analysis, we identify a key issue, \textit{i.e.}, \textbf{gradient entanglement}, which 1) distorts supervised gradients and weakens discrimination among known classes, and 2) induces representation-subspace overlap between known and novel classes, reducing the separability of novel categories. To address this issue, we propose the Energy-Aware Gradient Coordinator (EAGC), a plug-and-play gradient-level module that explicitly regulates the optimization process. EAGC comprises two components: Anchor-based Gradient Alignment (AGA) and Energy-aware Elastic Projection (EEP). AGA introduces a reference model to anchor the gradient directions of labeled samples, preserving the discriminative structure of known classes against the interference of unlabeled gradients. 
EEP softly projects unlabeled gradients onto the complement of the known-class subspace and derives an energy-based coefficient to adaptively scale the projection for each unlabeled sample according to its degree of alignment with the known subspace, thereby reducing subspace overlap without suppressing unlabeled samples that likely belong to known classes. 
Experiments show that EAGC consistently boosts existing methods and establishes new state-of-the-art results. Code is available at \url{https://haiyangzheng.github.io/EAGC}.
\end{abstract}    
\section{Introduction}
\label{sec:intro}

\begin{figure}[]
  \centering
  \includegraphics[width=\linewidth]{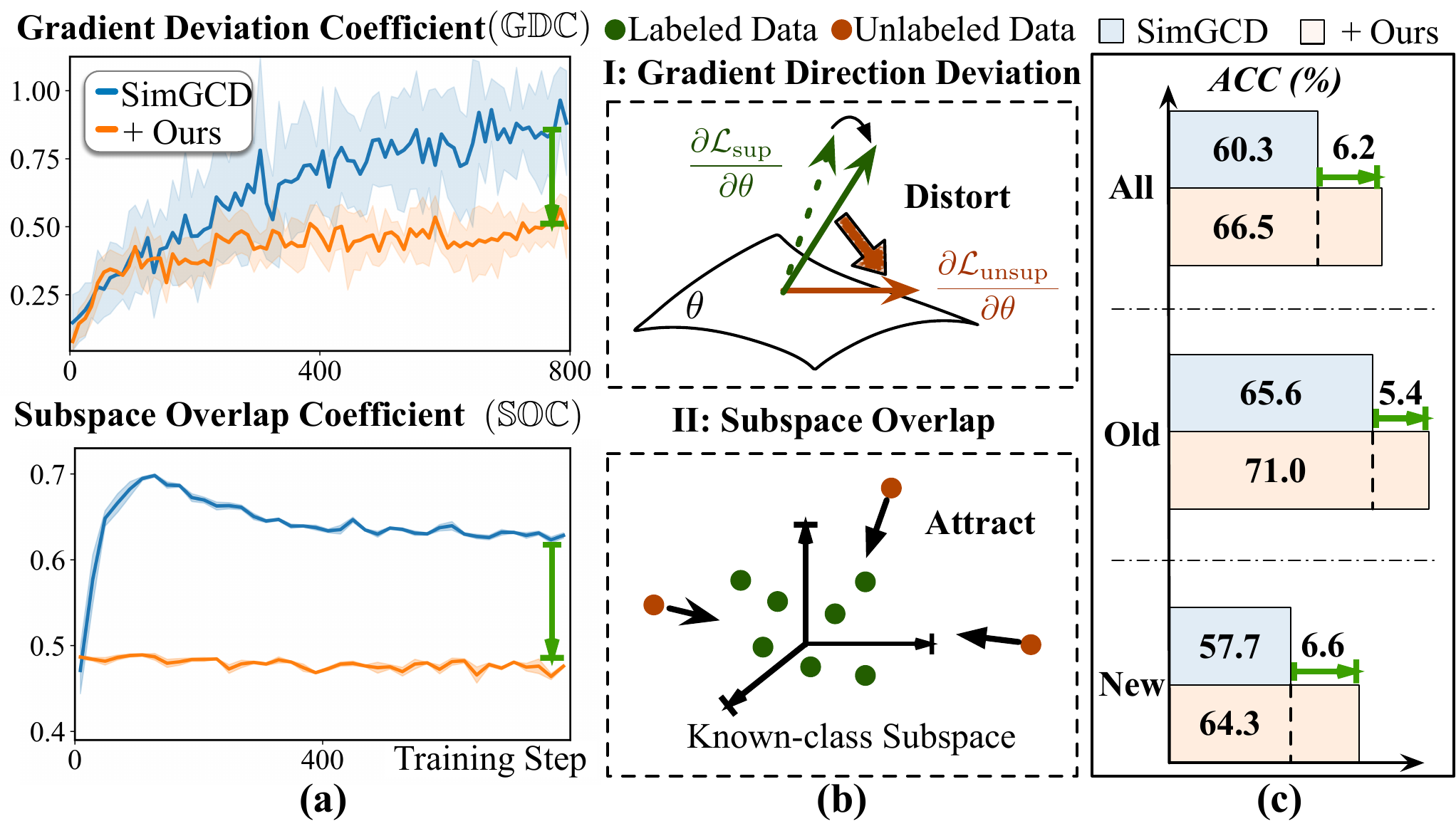}
    \caption{(a) $\mathbb{GDC}$ and $\mathbb{SOC}$ measure \textbf{Gradient Entanglement} (GE) from the gradient-direction and feature-subspace perspectives, respectively. Both indices reveal that existing GCD methods (e.g., SimGCD~\cite{simgcd}) suffer from non-negligible optimization conflicts. (b) We attribute this issue to two main factors. \textbf{I: Gradient Direction Deviation.} Unsupervised gradients aim to discover novel categories but are inherently noisy and unreliable, which \textbf{distort} the optimization direction dictated by the supervised objective that focuses on known categories. \textbf{II: Subspace Overlap.} Noisy and unreliable unsupervised objectives produce highly divergent gradients, whereas gradients of the supervised objective remain relatively consistent. Joint optimization inevitably lets supervised gradients dominate, which progressively \textbf{attract} novel-class representations toward the known-class subspace. (c) Our method effectively mitigates GE, yielding substantial performance gains.}
\label{fig:highlight}
\end{figure}
Category discovery—first introduced as Novel Category Discovery (NCD)~\cite{han2019learning} and later extended to Generalized Category Discovery (GCD)~\cite{gcd}—has recently emerged as an important problem in open-world visual learning. In GCD, a dataset is divided into a labeled subset containing known classes and an unlabeled subset that mixes known and novel categories. The goal is to leverage knowledge from the labeled subset to assign category labels to all samples in the unlabeled subset. Depending on how category labels are inferred, existing approaches can be broadly grouped into non-parametric methods~\cite{xcon,sun2022opencon,dccl,promptcal,infosieve,sun2024graph,gpc,cms,zheng2025generalized}, which perform clustering in the representation space, and parametric methods~\cite{ossl,yang2022comex,simgcd,pim,wang2023discover,ugcd,amend,liu2025debgcd,sptnet,legogcd,protogcd,peng2025mos,zhang2025less,he2025seal}, which learn explicit classifiers to jointly recognize known and novel categories.

\par
While existing GCD methods have achieved promising results, many rely on a direct combination of supervised and unsupervised objectives, which often leads to an \textit{unsatisfactory trade-off between known and novel classes}.
To investigate this issue in depth, we define two quantitative metrics: \textit{Gradient Deviation Coefficient (\(\mathbb{GDC}\))} and \textit{Subspace Overlap Coefficient (\(\mathbb{SOC}\))} (see details in~\cref{sec:preliminaries}). \(\mathbb{GDC}\) measures how unsupervised objectives interfere with the optimization direction of supervised gradients, whereas \(\mathbb{SOC}\) quantifies the degree of overlap between the representation subspaces of known and novel classes.
As shown in~\cref{fig:highlight}(a, b), we analyze the representative baseline SimGCD~\cite{simgcd} and find that: \textbf{Issue \ding{172}} The \(\mathbb{GDC}\) value gradually increases as training progresses, indicating that the optimization direction of supervised gradients is increasingly distorted by the unsupervised objectives, thereby weakening discrimination among known classes. \textbf{Issue \ding{173}} The \(\mathbb{SOC}\) rises sharply in the early stage and remains at a high level afterward, suggesting that novel-class representations are progressively drawn toward the known-class subspace, which further reduces their separability. We refer to these underexplored issues in GCD as \textbf{Gradient Entanglement (GE)}.

\par
To address these issues, we propose an Energy-Aware Gradient Coordinator (EAGC), which explicitly mitigates the \textbf{GE} problem from the perspective of gradient optimization, thereby consistently improving existing GCD methods (see~\cref{fig:highlight}(c)).
EAGC comprises two complementary components: Anchor-based Gradient Alignment (AGA) and Energy-aware Elastic Projection (EEP). To address \textbf{Issue \ding{172}}, AGA leverages a reference model trained under pure supervision as an anchor. It explicitly aligns the gradient direction of labeled samples toward this reference, thereby preserving the discriminative structure of known classes against interference from the unsupervised objectives. To mitigate \textbf{Issue \ding{173}}, EEP first constructs a Conceptor-based subspace of known classes and projects the gradients of unlabeled samples onto its complement. Considering the GCD characteristic that \textit{some unlabeled samples may also come from known categories}, EEP further introduces an energy-aware adaptive weighting strategy. This strategy determines the projection strength of each sample according to its feature energy ratio within the known-class subspace, preventing the over-suppression of known samples while effectively decoupling novel representations. \textbf{Notably}, EAGC operates in a plug-and-play fashion and integrates seamlessly with both parametric and non-parametric GCD frameworks without modifying the backbone architecture or the training objectives of the existing frameworks.

\par
Our main contributions are summarized as:
\begin{itemize}
    \item We quantitatively analyze the gradient behavior of existing GCD methods and identify a critical yet under-explored issue, \textit{i.e.}, \textbf{Gradient Entanglement (GE)}, which substantially constrains their performance.

    \item We propose a \textbf{plug-and-play} gradient-level coordinator, \textbf{EAGC}, which effectively mitigates GE and can be seamlessly integrated into existing GCD methods.

    \item We \textbf{derive} two task-specific modules from \textbf{theoretical analysis}. Inspired by proximal regularization, AGA primarily preserves the discrimination of known categories. EEP adaptively enhances the separability of novel categories, motivated by an energy-aware subspace analysis.

    \item Extensive experiments demonstrate that EAGC consistently enhances various parametric and non-parametric GCD methods, achieving new state-of-the-art performance across multiple benchmarks.

\end{itemize}

\section{Related Work}
\label{sec:related_work}
\paragraph{Category Discovery.}
Novel Category Discovery (NCD) was first introduced by Han et al.~\cite{han2019learning}, who formulated the problem as transferable clustering, \textit{i.e.}, transferring knowledge from labeled (seen) categories to cluster unlabeled (unseen) ones.
Since then, extensive efforts~\cite{QING202124,openmix2020,ncl,uno,dualRs,9747827,PSSCNCD,li2023closerlooknovelclass,ResTune,li2023supervised,Li2023ncdiic,sun2023nscl,peiyan2023class,wei2023ncdSkin,10328468,hasan2023novelcategoriesdiscoveryconstraints} have been devoted to improving representation learning and clustering reliability under this paradigm. To better reflect real-world scenarios, Generalized Category Discovery (GCD)~\cite{gcd} relaxes the NCD assumption by allowing unlabeled data to contain both known and unknown categories, inspiring a wide range of follow-up research. On one hand, several GCD studies investigate more practical settings, such as federated GCD~\cite{fgcd,zhang2023unbiasedtrainingfederatedopenworld,wang2023federatedcontinualnovelclass}, on-the-fly discovery~\cite{ocd,phe,liu2025generate}, ultra-fine category discovery~\cite{ultrafine}, multi-modal extensions~\cite{clipgcd,textgcd,mgcd,get,yang2025consistent}, and domain-shift scenarios~\cite{Yu_Ikami_Irie_Aizawa_2022,zhuang2022opensetdomainadaptation,zang2023boostingnovelcategorydiscovery,rongali2024cdadnetbridgingdomaingaps,wang2024exclusivestyleremovalcross,wang2024hilo}, reflecting the growing interest in open-world recognition. 
\par
On the other hand, a large body of GCD work focuses on improving generalization, calibration, and clustering robustness within the standard GCD setting.
Existing approaches can be broadly divided into \textbf{parametric} methods~\cite{ossl,yang2022comex,simgcd,pim,wang2023discover,ugcd,amend,liu2025debgcd,sptnet,legogcd,protogcd,peng2025mos,zhang2025less,he2025seal}, which rely on a classifier with a predefined number of classes, and \textbf{non-parametric} methods~\cite{xcon,sun2022opencon,dccl,promptcal,infosieve,sun2024graph,gpc,cms,zheng2025generalized}, which eliminate this constraint.
Moreover, recent studies~\cite{aplgcd,congcd,hypcd} propose plug-and-play modules that can be seamlessly integrated into existing frameworks to further enhance discovery performance. Following this line, our EAGC is also designed as a plug-and-play algorithm. Moreover, \textit{distinct from prior efforts, EAGC analyzes the GCD challenge from an \textbf{optimization perspective} and identifies the inherent Gradient Entanglement issue. By effectively mitigating this problem, EAGC improves existing methods under both parametric and non-parametric paradigms, achieving new state-of-the-art performance.}

\paragraph{Gradient Projection.}
Gradient projection is a technique designed to regulate the direction of parameter updates by projecting gradients onto constrained subspaces, thereby mitigating interference between tasks. This approach has been widely applied in multi‑task learning~\cite{Yu2020GradientSurgery,huang2025directional} (MTL) and continual learning~\cite{Saha2021GPM} (CL) to reduce gradient conflicts and catastrophic forgetting. For instance, in MTL, Gradient Surgery~\cite{Yu2020GradientSurgery} projects a given task’s gradient onto the normal plane of any conflicting task gradient, thus alleviating harmful interference and improving overall optimization. In CL, GPM~\cite{Saha2021GPM} uses singular value decomposition on the task‑activations to identify a core gradient subspace, and then during subsequent task training forces new‑task gradients to be orthogonal to that subspace—thereby preserving old‑task knowledge. Additionally, PGP~\cite{Qiao2024PGP} integrates prompt‑tuning with gradient projection, demonstrating that by ensuring prompt‑gradient updates are orthogonal to previous feature subspaces, one can reduce forgetting while using parameter‑efficient adaptation. \textit{Unlike prior methods designed for fully supervised settings, our EAGC is \textbf{tailored} for the GCD setting, where 1) the model must learn jointly from both labeled and \textbf{unlabeled data}, and 2) the unlabeled subset uniquely contains not only known classes but also \textbf{novel ones}. These characteristics give rise to GCD-specific \textbf{gradient entanglement} issues.}

\section{Preliminaries}
\label{sec:preliminaries}

\noindent\textbf{Problem Setup.} In GCD, we are given a labeled dataset 
$\mathcal{D}_{L} = \left\{ (\mathbf{x}^{l}_{i}, y^{l}_{i}) \mid (\mathbf{x}^{l}_{i}, y^{l}_{i}) \in \mathcal{X} \times \mathcal{Y}_{L} \right\}_{i=1}^{N}$ and an unlabeled dataset 
$\mathcal{D}_{U} = \left\{ \mathbf{x}^{u}_{i} \mid \mathbf{x}^{u}_{i} \in \mathcal{X} \right\}_{i=1}^{M}$. 
Here, $N$ and $M$ denote the numbers of labeled and unlabeled samples, respectively, 
and $\mathcal{Y}_{L}$ represents the label space of known classes. 
The unlabeled dataset is drawn from a broader label space $\mathcal{Y}_{U}$, 
which contains both known and unknown categories, i.e., $\mathcal{Y}_{L} \subset \mathcal{Y}_{U}$. 
The goal of GCD is to automatically categorize unlabeled samples, 
distinguishing between known (old) and novel (unknown) classes by leveraging 
the knowledge learned from labeled data.

\subsection{Gradient Entanglement in GCD}
\paragraph{Definition.}
The overall learning objective in GCD is typically expressed as:
\begin{equation}
\mathcal{L}_{\text{GCD}} = \alpha \mathcal{L}_{\text{sup}}(\mathcal{D}_L) + \beta \mathcal{L}_{\text{unsup}}(\mathcal{D}_U),
\label{eq:gcd_objective}
\end{equation}
where $\mathcal{L}_{\text{sup}}$ provides reliable supervision from labeled data, while $\mathcal{L}_{\text{unsup}}$ relies on noisy self-supervised or pseudo-label signals—particularly unreliable in the early stage of training.
This supervision asymmetry can lead to two coupled effects: \textit{(1) unreliable unsupervised gradients interfere with the optimization of $\mathcal{L}_{\text{sup}}$, and (2) dominant supervised gradients pull unlabeled representations toward the known-class subspace.}
From the perspective of gradient optimization, we refer to this phenomenon as \textit{Gradient Entanglement} (\textbf{GE}). Specifically, the gradient of $\mathcal{L}_{\text{GCD}}$ with respect to model parameters $\theta$ can be decomposed as:
\begin{equation}
g = \nabla_\theta \mathcal{L}_{\text{GCD}}= g_L + g_U, g_L = \alpha \nabla_\theta \mathcal{L}_{\text{sup}}, g_U = \beta \nabla_\theta \mathcal{L}_{\text{unsup}}.
\label{eq:grad}
\end{equation}
\textit{First}, the unsupervised gradient $g_U$ may conflict with or distort the optimization direction of the supervised gradient $g_L$, thereby weakening the discriminative power of known classes and leading to suboptimal parameter updates.
\textit{Second}, since known classes contribute gradients from both labeled supervision ($g_L$) and the known-class portion of the unlabeled data ($g_U^{\text{known}}$), these gradients tend to dominate the optimization dynamics and pull novel representations toward the known-class subspace, thereby reducing the inter-class separability of novel categories.


\noindent\textbf{Observation~\ding{172}. Gradient Direction Deviation.} 
Let $\hat{g_L} = \nabla_\theta \mathcal{L}_{\text{sup}}$ denote the reference gradient obtained when optimizing only $\mathcal{L}_{\text{sup}}$ (i.e., setting $\beta = 0$ in \cref{eq:gcd_objective}).  
To quantify the effect of unlabeled data on the supervised optimization dynamics, we define the \textit{Gradient Deviation Coefficient (\(\mathbb{GDC}\))} as:
\begin{equation}
    \mathbb{GDC} = 1 - \frac{\langle \hat{g_L}, g \rangle}{\|\hat{g_L}\| \, \|g\|}.
\end{equation}
It ranges within $[0,2]$ and quantifies how unlabeled data interferes with the supervised optimization signal—larger values indicate a more severe GE issue.

\noindent\textbf{Observation~\ding{173}. Subspace Overlap.} 
Let the feature matrix of labeled (known-class) data be $Z_{\text{old}} \in \mathbb{R}^{N \times d}$ and that of novel classes (from unlabeled data) be $Z_{\text{new}} \in \mathbb{R}^{M_n \times d}$, where $M_n$ denotes the number of novel samples and $d$ is the feature dimension. Let $U_k$ denote the top-$k$ principal components of $Z_{\text{old}}$, and define the projection matrix as $P_{\text{old}} = U_k U_k^{\top}$, which captures the dominant energy directions of known classes in the feature space. 
To measure the representation subspace overlap between novel and known classes, we define the \textit{Subspace Overlap Coefficient (\(\mathbb{SOC}\))} as:
\begin{equation}
\mathbb{SOC} = \frac{\| Z_{\text{new}} P_{\text{old}} \|_F^2}{\| Z_{\text{new}} \|_F^2}.
\end{equation}
It ranges within $[0,1]$ and quantifies the proportion of novel-class feature energy projected onto the subspace of known classes—a higher \(\mathbb{SOC}\) indicates a stronger tendency of novel features to align with the known-class subspace.
\begin{figure*}[t]
  \centering
  \includegraphics[width=\linewidth]{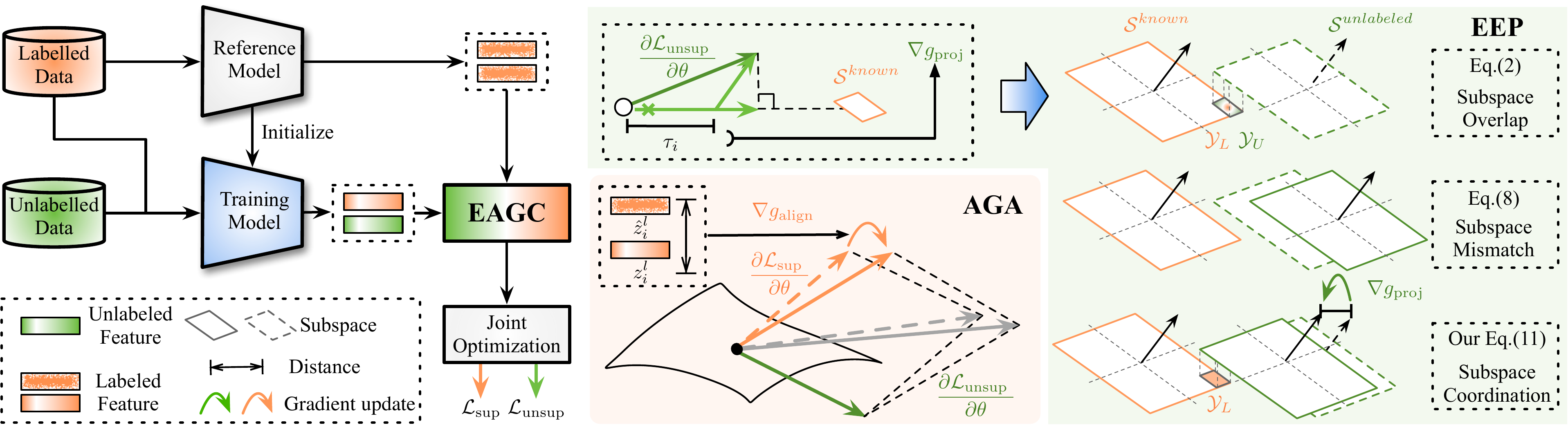}
  
\caption{Framework Overview of the proposed Energy-Aware Gradient Coordinator (\textbf{EAGC}).
EAGC consists of two components: (1) \textit{Anchor-based Gradient Alignment} (\textbf{AGA}), which uses feature anchors from a reference model to generate the alignment term $\nabla g_{\text{align}}$ to stabilize the updates of $\mathcal{L}_{\text{sup}}$, and (2)~\textit{Energy-aware Elastic Projection} (\textbf{EEP}), which generates the projection term $\nabla g_{\text{proj}}$ by projecting the unlabeled-gradient component $\nabla \mathcal{L}_{\text{unsup}}$ onto the complement of the known-class subspace $\mathcal{S}^{\text{known}}$ using a sample-adaptive elastic weight~$\tau$.
}
    \label{fig:framework}
\end{figure*}

\section{Method}
\noindent\textbf{Motivation.}
To address Gradient Entanglement, our motivation is twofold. 
First, to stabilize the optimization of labeled samples against disturbances from unlabeled objectives, we seek to \textit{anchor the gradient direction of labeled data} and maintain a consistent optimization trajectory for known classes. 
Second, to mitigate the attraction of novel representations toward the known-class subspace, we further aim to \textit{project the gradient direction of unlabeled data} away from the dominant directions of known categories, thereby enhancing their separability.

\noindent \textbf{Framework Overview.}
We propose an Energy-Aware Gradient Coordinator (EAGC) module to explicitly mitigate Gradient Entanglement in GCD from the perspective of gradient optimization. 
As illustrated in \cref{fig:framework}, EAGC comprises two complementary components designed to address the two manifestations of Gradient Entanglement. 
(1)~\textit{Anchor-based Gradient Alignment (AGA)} focuses on stabilizing the optimization of labeled data. 
It anchors the gradient direction of labeled samples toward a reference model trained purely on labeled data, thereby preserving stable supervision against disturbances from unlabeled objectives. 
(2)~\textit{Energy-aware Elastic Projection (EEP)} operates on unlabeled data. 
It performs an elastic soft projection of unlabeled gradients away from the known-class subspace, with an adaptive weighting mechanism that prevents over-suppression of known samples while improving the separability of novel ones.  
Together, these two components act in a complementary manner—AGA preserves the discriminative boundary of known classes, while EEP encourages novel representations to evolve in a disentangled subspace, resulting in more stable and balanced GCD optimization.

\noindent \textbf{Notation.}
During backpropagation, the parameter gradient $\nabla_\theta \mathcal{L}$ is computed via the chain rule,
$\nabla_\theta \mathcal{L} = \frac{\partial z}{\partial \theta} \cdot \nabla_z \mathcal{L}$,  
where $\nabla_z \mathcal{L} = \frac{\partial \mathcal{L}}{\partial z}$ denotes the gradient of the loss with respect to the intermediate feature representation $z$.  
In our EAGC, we regulate gradient alignment and projection by modulating this feature-level gradient $\nabla_z \mathcal{L}$ before it is propagated back to the encoder parameters.

\subsection{Anchor-based Gradient Alignment} 
To mitigate the interference of unlabeled objectives on the optimization direction of labeled samples during joint training, 
we introduce a gradient alignment constraint that maintains stable and discriminative learning for known classes.

\noindent\textbf{Reference Model.} 
Given the labeled subset $\mathcal{D}_L$, we first train a reliable \textit{reference model} using the supervised classification objective:
\begin{equation}
\mathcal{L}_{\text{cls}} = \frac{1}{|\mathcal{B}_L|} \sum_{i \in \mathcal{B}_L} 
\ell(y_i^l, p_i),
\label{loss_cls_sup}
\end{equation}
where $\mathcal{B}_L$ denotes a mini-batch from $\mathcal{D}_L$, 
$\ell(\cdot)$ is the cross-entropy loss, 
$p_i$ is the predicted probability distribution of the labeled input $x_i^l$, 
and $y_i^l$ is its corresponding label. 
Specifically, $p_i = \sigma(f(\mathcal{E}(x_i^l))/\tau_s)$, 
where $\mathcal{E}(\cdot)$ denotes the image encoder, 
$f(\cdot)$ is a parametric classifier, 
$\sigma$ is the softmax function, and $\tau_s$ is a temperature parameter. 
The resulting \textit{reference model}, denoted as $\mathcal{E}_r(\cdot)$, 
is fixed during GCD training to provide stable feature anchors for known classes.

\noindent\textbf{Gradient Alignment.}
During GCD optimization with the objective $\mathcal{L}_{\text{GCD}}$~(\cref{eq:gcd_objective}), 
we impose a gradient alignment constraint on labeled samples:
\begin{equation}
\nabla_{z} \mathcal{L}_{\text{GCD}} 
= \nabla_{z^l} \mathcal{L}_{\text{sup}} 
+ \underbrace{\nabla_{z^u} \mathcal{L}_{\text{unsup}}}_{\text{Disturbance}}
+ \underbrace{\lambda_a (z^l - \hat{z}^l)}_{\text{Stabilization}},
\end{equation}
where $z^l$ denotes the feature representations of labeled samples, 
$z^l_i = \mathcal{E}(x_i^l)$, and $\hat{z}^l_i = \mathcal{E}_r(x_i^l)$. 
Here, $\nabla_{z^l} \mathcal{L} = \frac{\partial \mathcal{L}}{\partial z^l}$ represents the gradient of the loss function with respect to $z^l$, 
and $\lambda_a$ controls the strength of the alignment correction. 
The alignment term $\nabla_{z^l} g_{\text{align}} = \lambda_a (z^l - \hat{z}^l)$ 
anchors the gradient direction of labeled samples toward that of the reference model, 
thereby stabilizing the optimization against disturbances introduced by $\mathcal{L}_{\text{unsup}}$.

\noindent\textbf{Remark.}
The stabilization term can be interpreted as a form of proximal regularization~\cite{nocedal2006numerical,yun2021adaptive} in the feature space, 
which effectively defines a local trust region around the supervised optimum $\hat{z}$. 
This induces a contraction effect that stabilizes the representations of labeled samples and mitigates semantic drift. 
From the objective perspective, we further verify that such proximal regularization consistently enhances GCD performance, as reported in Appendix~C.


\subsection{Energy-aware Elastic Projection}
During joint optimization, the strong supervision from labeled data tends to generate dominant gradients that steer learning toward known categories, consequently pulling novel representations into the known-class subspace and hindering the formation of distinct clusters. To mitigate this subspace overlap, we explicitly project unlabeled gradients away from the known-class subspace.

\noindent\textbf{Known-Class Subspace Construction.}
Our goal is to construct a representation subspace for known classes such that novel representations can be projected into regions of the feature space not dominated by known categories. 
While PCA provides a hard low-rank approximation by retaining only the principal energy directions, it tends to over-compress the feature space and discard low-energy components that may encode subtle semantic variations distinguishing known and novel classes. 
To address this, we adopt a Conceptor-based~\cite{jaeger2014conceptor} representation subspace that models the principal directions of known classes in an energy-weighted and flexible manner. 
Given the feature matrix of the labeled subset 
$Z_{\text{old}} = \{\mathcal{E}_r(\mathbf{x}^l_i)\} \in \mathbb{R}^{N \times d}$, 
the known-class subspace is computed as:
\begin{equation}
\mathcal S^{known} = R (R + \eta^{-2} I)^{-1}, \quad
R = \frac{1}{N} Z_{\text{old}}^\top Z_{\text{old}},
\label{eq:conceptor}
\end{equation}
where $\eta$ is the \textit{aperture} controlling the softness of subspace capture and $I$ is the identity matrix. 
The resulting $\mathcal S^{known}$ serves as a soft projection operator that preserves the dominant energy directions of known classes. 
Given a feature vector $z_i$, its projected component within the known-class subspace is $z_{\text{proj}} = z_i \mathcal S^{known}$.

\noindent\textbf{Gradient Projection.}
For unlabeled samples, we regulate the feature-level gradients to reduce their overlap with the known-class subspace. 
Let $\nabla_{z^u} \mathcal{L}_{\text{unsup}}$ denote the gradient of the unsupervised loss with respect to unlabeled feature representations $z^u$. 
We apply a soft projection defined as:
\begin{equation}
\nabla_{z^u} \mathcal{L}_{\text{unsup}}
= \nabla_{z^u} \mathcal{L}_{\text{unsup}}
- \lambda_p \big(\nabla_{z^u} \mathcal{L}_{\text{unsup}} \mathcal S^{known}\big),
\end{equation}
where $\lambda_p$ controls the overall projection strength. 
However, since the unlabeled set also contains samples from known categories, 
projecting all unlabeled gradients away from the known-class subspace would undesirably \textit{suppress the learning of known-category samples}.

\noindent\textbf{Energy-Aware Adaptive Weighting.}
To address this issue, we introduce a sample-specific projection weight determined by each feature’s energy ratio within the known-class subspace:
\begin{equation}
\mathbf{E}_{\text{old}}(z_i)
= \frac{z_i^{\top} \mathcal S^{known} z_i}{\|z_i\|_2^2}
\in [0,1],
\end{equation}
which measures the proportion of the feature’s squared magnitude (i.e., feature energy) residing in the known-class subspace.
We then compute the average energy ratio of labeled data, $\overline{\mathbf{E}_{\text{old}}^l}$, and define the adaptive projection weight for each unlabeled feature as:
\begin{equation}
\tau_i = 1 - \frac{\mathbf{E}_{\text{old}}(z_i^u)}{\overline{\mathbf{E}_{\text{old}}^l}}.
\label{eq:proj_weight}
\end{equation}
This normalization calibrates the projection strength relative to the alignment of labeled samples: 
features that are highly aligned with the known subspace (likely known) receive weaker projection, 
whereas those less aligned (likely novel) are projected more strongly.
The resulting Elastic Gradient Projection is thus formulated as:
\begin{equation}
\nabla_{z^u} g_{\text{proj}} = - \lambda_p \big(\tau \cdot \nabla_{z^u} \mathcal{L}_{\text{unsup}} \mathcal S^{known}\big).
\end{equation}

\noindent\textbf{Remark.}
Our design exhibits two key properties.
(1) The known-class representation subspace is constructed using Conceptor theory~\cite{jaeger2014conceptor}, which provides a soft, energy-weighted characterization of dominant feature directions. 
Unlike hard low-rank truncations such as PCA, this formulation offers greater flexibility in modeling the boundary between known and novel categories. 
(2) Because the unlabeled set contains both known and novel samples, 
we employ an \textit{Energy-aware Elastic Projection} that adaptively modulates the projection strength based on each sample’s alignment with the known-class subspace. 
This sample-specific adjustment avoids over-suppressing known-category gradients while effectively decoupling novel representations.

\subsection{Plug-and-Play Training}
We integrate the proposed EAGC module as a plug-and-play component that operates purely at the gradient level, 
requiring no modification to either the network architecture or the original training objectives. 
Under EAGC, the overall feature-level gradient update can be formulated as:
\begin{equation}
\nabla_z \mathcal{L}_{\text{GCD}} 
= \nabla_z \mathcal{L}_{\text{sup}} 
+ \nabla_z \mathcal{L}_{\text{unsup}} 
+ \nabla_{z^l} g_{\text{align}} 
+ \nabla_{z^u} g_{\text{proj}}.
\end{equation}
The complete training pipeline of EAGC is summarized in Algorithm~\ref{alg:eagc}.

\section{Experiments}

\subsection{Experimental Setup}

\begin{algorithm}[h]
\caption{Energy-Aware Gradient Coordinator}
\label{alg:eagc}
\SetAlgoLined
\KwIn{
Labeled data $\mathcal{D}_L$, unlabeled data $\mathcal{D}_U$, image encoder $\mathcal{E}(\cdot)$, 
projection head $\phi(\cdot)$, hyperparameters $(\lambda_a, \lambda_p, \eta)$.
}

/* \textit{Getting reference model} */ \\
Train $\mathcal{E}(\cdot)$ on $\mathcal{D}_L$ using $\mathcal{L}_{cls}$ in \cref{loss_cls_sup} to obtain $\mathcal{E}_r$\;
Freeze $\mathcal{E}_r$ to serve as an anchor for labeled data\;

\BlankLine
/* \textit{Known-class Subspace Construction} */ \\
Compute $Z_{\text{old}}$ and $\mathcal S^{known}$ using \cref{eq:conceptor}\;
Compute the average energy ratio $\overline{\mathbf{E}_{\text{old}}^l}$\;

\BlankLine
/* \textit{GCD Training with EAGC} */ \\
\For{$epoch = 1$ \textbf{to} $T$}{
    \For{batch $({x}, y, mask_{\text{lab}})$ from $\mathcal{D}_L \cup \mathcal{D}_U$}{
        /* \textit{Anchor-based Gradient Alignment} */
        $z=\mathcal E(x), \hat{z}=\mathcal{E}_r(x)$\;
        Register backward hook for alignment term: 
        $mask_{\text{lab}} \cdot \lambda_a (z - \hat{z})$\;

        \BlankLine
        /* \textit{Forward pass} */ \\
        Obtain $h=\phi(z)$ and compute 
        $\mathcal{L}_{\text{GCD}} = \alpha\mathcal{L}_{\text{sup}} + \beta\mathcal{L}_{\text{unsup}}$\;

        \BlankLine
        /* \textit{Energy-aware Gradient Projection} */ \\
        Compute projection weights $\tau$ via \cref{eq:proj_weight}\;
        Register backward hook for projection term: 
        $-(1-mask_{\text{lab}})\cdot \lambda_p (\tau \cdot \nabla_z \mathcal{L}_{\text{unsup}} \mathcal S^{known})$\;

        \BlankLine
        /* \textit{Backward Pass} */ \\
        Update $\mathcal{E}(\cdot)$ and $\phi(\cdot)$ via an optimizer step\;
    }
}
\end{algorithm}

\noindent\textbf{Datasets.} 
We evaluate our method across multiple datasets, including two generic benchmarks, CIFAR-100~\cite{cifar} and ImageNet-100~\cite{imagenet}, as well as the Semantic Shift Benchmark (SSB)~\cite{ssb}, which comprises three fine-grained datasets: CUB-200-2011~\cite{cub}, Stanford Cars~\cite{scars}, and FGVC-Aircraft~\cite{aircraft}. 
Following the protocol of~\cite{gcd}, a subset of categories is designated as the labeled set $\mathcal{Y}_L$. 
Within these classes, 50\% of the images are used to form the labeled subset $\mathcal{D}_L$, while the remaining images constitute the unlabeled subset $\mathcal{D}_U$. 
Dataset statistics are provided in Appendix~A.

\noindent\textbf{Evaluation Metric.} 
Following~\cite{gcd}, we report clustering accuracy (ACC), where predicted clusters are matched to ground-truth labels using the Hungarian algorithm.

\noindent\textbf{Implementation Details.} 
We evaluate our EAGC module on both parametric baselines (SimGCD~\cite{simgcd}, LegoGCD~\cite{legogcd}, and SPTNet~\cite{sptnet}) and the non-parametric baseline SelEx~\cite{selex}. 
Unless otherwise specified, experiments adopt a ViT-B/16 backbone pretrained with DINO~\cite{dino}. 
All baselines are trained for 200 epochs with a batch size of 128. The learning rate is set to 0.1 for the backbone and 1.0 for the projection head, both scheduled by cosine decay. 
The reference model follows the same trainable-layer configuration as the baselines and is trained for 30 epochs (only 3 for CIFAR-100 to prevent overfitting on known classes due to its 80/20 split) with $\tau_s$ = 0.1 and a cosine-decayed learning rate of 0.02.
Other training hyperparameters follow the default settings of the respective baselines. 
Our EAGC introduces an additional hyperparameter $\eta$ = 2.0, which controls the aperture of the conceptor.
For all baselines and datasets, we uniformly set $(\lambda_a, \lambda_p)$ = (0.7, 0.5), which are tuned using SimGCD on the CUB dataset.
All experiments are conducted on NVIDIA 3090Ti GPUs, and results are averaged over three runs with different random seeds.

\begin{table*}[!h]
\centering
\caption{Comparison with state-of-the-art GCD methods. Best and second-best results are \textbf{bold} and \underline{underlined}, respectively.}

\setlength{\tabcolsep}{2.5pt}
\renewcommand{\arraystretch}{0.9}
\begin{tabular}{l|ccc|ccc|ccc|ccc|ccc|ccc}
\toprule
\multirow{2}{*}{Method}
& \multicolumn{3}{c|}{CUB}
& \multicolumn{3}{c|}{Stanford-Cars}
& \multicolumn{3}{c|}{Aircraft}
& \multicolumn{3}{c|}{CIFAR-100}
& \multicolumn{3}{c|}{ImageNet-100}
& \multicolumn{3}{c}{Average} \\
\cmidrule(lr){2-4}\cmidrule(lr){5-7}\cmidrule(lr){8-10}\cmidrule(lr){11-13}\cmidrule(lr){14-16}\cmidrule(lr){17-19}
& All & Old & New & All & Old & New & All & Old & New & All & Old & New & All & Old & New & All & Old & New \\
\midrule
ORCA~\cite{ossl}
&36.3&43.8&32.6 
&31.6&32.0&31.4 
&31.9&42.2&26.9  
&73.5&\textbf{92.6}&63.9 
&81.8&86.2&79.6
&51.0&59.4&46.9  \\

GCD~\cite{gcd}
& 51.3 & 56.6 & 48.7
& 39.0 & 57.6 & 29.9
& 45.0 & 41.1 & 46.9
& 73.0 & 76.2 & 66.5
& 74.1 & 89.8 & 66.3
& 56.5 & 64.3 & 51.7 \\

GPC~\cite{gpc}
& 52.0 & 55.5 & 47.5
& 38.2 & 58.9 & 27.4
& 43.3 & 40.7 & 44.8
& 75.4 & 84.6 & 60.1
& 75.3 & 93.4 & 66.7
& 56.8 & 66.6 & 49.3  \\

XCon~\cite{xcon}
& 52.1 & 54.3 & 51.0
& 40.5 & 58.8 & 31.7
& 47.7 & 44.4 & 49.4
& 74.2 & 81.2 & 60.3
& 77.6 & 93.5 & 69.7
& 58.4 & 66.4 & 52.4 \\

PromptCAL~\cite{promptcal}
& 62.9 & 64.4 & 62.1
& 50.2 & 70.1 & 40.6
& 52.2 & 52.2 & 52.3
& 81.2 & 84.2 & 75.3
& 83.1 & 92.7 & 78.3
& 65.9 & 72.7 & 61.7 \\

DCCL~\cite{dccl}
& 63.5 & 60.8 & 64.9 
& 43.1 & 55.7 & 36.2 
& -- & -- & -- 
& 75.3 & 76.8 & 70.2 
& 80.5 & 90.5 & 76.2
& -- & -- & --\\

$\mu$GCD~\cite{ugcd}
& 65.7 & 68.0 & 64.6
& 56.5 & 68.1 & 50.9
& 53.8 & 55.4 & 53.0
& -- & -- & --
& -- & -- & --
& -- & -- & -- \\

InfoSieve~\cite{infosieve}
& 69.4 & \textbf{77.9} & 65.2
& 55.7 & 74.8 & 46.4
& 56.3 & 63.7 & 52.5
& 78.3 & 82.2 & 70.5
& 80.5 & 93.8 & 73.8
& 68.0 & 78.5 & 61.7 \\

AMEND~\cite{amend}
& 64.9 & 75.6 & 59.6
& 56.4 & 73.3 & 48.2
& 52.8 & 61.8 & 48.3
& 81.0 & 79.9 & \underline{83.3} 
& 83.2 & 92.9 & 78.3
& 67.7 & 76.7 & 63.5 \\

CMS~\cite{cms}
& 68.2 & \underline{76.5} & 64.0
& 56.9 & 76.1 & 47.6
& 56.0 & 63.4 & 52.3
& 82.3 & \underline{85.7} & 75.5 
& 84.7 & \underline{95.6} & 79.2
& 69.6 & \underline{79.5} & 63.7 \\

\cmidrule(lr){1-19}
SimGCD~\cite{simgcd}
& 60.3 & 65.6 & 57.7
& 53.8 & 71.9 & 45.0
& 54.2 & 59.1 & 51.8
& 80.1 & 81.2 & 77.8
& 83.0 & 93.1 & 77.9
& 66.3 & 74.2 & 62.0 \\

\cellcolor{my_blue}\quad + EAGC
& \cellcolor{my_blue}66.5 & \cellcolor{my_blue}71.0 & \cellcolor{my_blue}64.3
& \cellcolor{my_blue}\underline{62.9} & \cellcolor{my_blue}76.0 & \cellcolor{my_blue}\underline{56.6}
& \cellcolor{my_blue}57.7 & \cellcolor{my_blue}60.4 & \cellcolor{my_blue}56.3
& \cellcolor{my_blue}83.1 & \cellcolor{my_blue}84.1 & \cellcolor{my_blue}81.0
& \cellcolor{my_blue}83.5 & \cellcolor{my_blue}93.9 & \cellcolor{my_blue}78.3
&\cellcolor{my_blue}70.7&\cellcolor{my_blue}77.1&\cellcolor{my_blue}67.3\\


LegoGCD~\cite{legogcd}
& 63.8 & 71.9 & 59.8
& 57.3 & 75.7 & 48.4
& 55.0 & 61.5 & 51.7
& 81.8 & 81.4 & 82.5
& \underline{86.3} & 94.5 & \underline{82.1}
& 68.8 & 77.0 & 64.9 \\

\cellcolor{my_blue}\quad + EAGC
& \cellcolor{my_blue}64.6 & \cellcolor{my_blue}72.5 & \cellcolor{my_blue}60.7
& \cellcolor{my_blue}62.8 & \cellcolor{my_blue}76.2 & \cellcolor{my_blue}56.3
& \cellcolor{my_blue}56.7 & \cellcolor{my_blue}61.9 & \cellcolor{my_blue}54.0
& \cellcolor{my_blue}\underline{83.5} & \cellcolor{my_blue}83.7 & \cellcolor{my_blue}83.0
& \cellcolor{my_blue}84.8 & \cellcolor{my_blue} 94.4 & \cellcolor{my_blue}80.1
&\cellcolor{my_blue}70.5&\cellcolor{my_blue}77.7&\cellcolor{my_blue}66.8 \\


SPTNet~\cite{sptnet}
& 65.8 & 68.8 & 65.1
& 59.0 & \underline{79.2} & 49.3
& \underline{59.3} & 61.8 & \underline{58.1}
& 81.3 & 84.3 & 75.6
& 85.4 & 93.2 & 81.4
& 70.2 & 77.5 & 65.9 \\

\cellcolor{my_blue}\quad + EAGC
& \cellcolor{my_blue}67.4 & \cellcolor{my_blue}70.5 & \cellcolor{my_blue}66.0
& \cellcolor{my_blue}62.1 & \cellcolor{my_blue}76.1 & \cellcolor{my_blue}55.4
& \cellcolor{my_blue}57.0 & \cellcolor{my_blue}58.1 & \cellcolor{my_blue}56.5
& \cellcolor{my_blue}\textbf{84.0} & \cellcolor{my_blue}84.3 & \cellcolor{my_blue}\textbf{83.5}
& \cellcolor{my_blue}85.3 & \cellcolor{my_blue}94.3 & \cellcolor{my_blue}80.4
&\cellcolor{my_blue}\underline{71.2} &\cellcolor{my_blue}76.7 &\cellcolor{my_blue}\underline{68.4} \\


SelEx~\cite{selex}
& \underline{73.6} & 75.3 & \underline{72.8}
& 58.5 & 75.6 & 50.3
& 57.1 & \underline{64.7} & 53.3
& 82.3 & 85.3 & 76.3
& 83.1 & 93.6 & 77.8
&70.9&78.9&66.1 \\

\cellcolor{my_blue}\quad + EAGC
& \cellcolor{my_blue}\textbf{83.2} & \cellcolor{my_blue}73.9 & \cellcolor{my_blue}\textbf{87.9}
& \cellcolor{my_blue}\textbf{65.7} & \cellcolor{my_blue}\textbf{83.7} & \cellcolor{my_blue}\textbf{57.0}
& \cellcolor{my_blue}\textbf{65.7} & \cellcolor{my_blue}\textbf{65.6 }& \cellcolor{my_blue}\textbf{65.7}
& \cellcolor{my_blue}79.3 & \cellcolor{my_blue}84.1 & \cellcolor{my_blue}69.7
& \cellcolor{my_blue}\textbf{89.8} & \cellcolor{my_blue}\textbf{96.2} & \cellcolor{my_blue}\textbf{86.5}
&\cellcolor{my_blue}\textbf{76.7}&\cellcolor{my_blue}\textbf{80.7}&\cellcolor{my_blue}\textbf{73.4} \\


\textbf{Avg. $\triangle$} 
&\textcolor{ForestGreen}{+4.6}&\textcolor{ForestGreen}{+1.6}&\textcolor{ForestGreen}{+5.9}
&\textcolor{ForestGreen}{+6.2}&\textcolor{ForestGreen}{+2.4}&\textcolor{ForestGreen}{+8.1}
&\textcolor{ForestGreen}{+2.9}&\textcolor{red!50!white}{-0.3}&\textcolor{ForestGreen}{+4.4}
&\textcolor{ForestGreen}{+1.1}&\textcolor{ForestGreen}{+1.0}&\textcolor{ForestGreen}{+1.3}
&\textcolor{ForestGreen}{+1.4}&\textcolor{ForestGreen}{+1.1}&\textcolor{ForestGreen}{+1.5}
&\textcolor{ForestGreen}{+3.2}&\textcolor{ForestGreen}{+1.2}&\textcolor{ForestGreen}{+4.3}
\\

\bottomrule
\end{tabular}

\label{tab:sota}
\end{table*}

\subsection{Quantitative Comparison}
We compare our method against a broad range of GCD approaches, including ORCA~\cite{ossl}, GCD~\cite{gcd}, GPC~\cite{gpc}, XCon~\cite{xcon}, PromptCAL~\cite{promptcal}, DCCL~\cite{dccl}, $\mu$GCD~\cite{ugcd}, InfoSieve~\cite{infosieve}, AMEND~\cite{amend}, and CMS~\cite{cms}, as well as baselines SimGCD~\cite{simgcd}, LegoGCD~\cite{legogcd}, SPTNet~\cite{sptnet}, and SelEx~\cite{selex}. 
The results are reported in \cref{tab:sota}. 
Across four baselines and five datasets, integrating EAGC improves performance overall, yielding average gains of 3.2\% in All ACC and 4.3\% in New ACC.
Notably, when built upon SelEx, EAGC yields substantial improvements of 9.6\% and 6.7\% in All ACC on CUB and ImageNet-100, respectively.

\noindent\textbf{Results on Fine-grained Datasets.} 
The results on fine-grained datasets are presented on the left side of \cref{tab:sota}. 
Across the three fine-grained benchmarks, incorporating EAGC improves category discovery accuracy overall.
On average across the three fine-grained benchmarks, EAGC brings gains of 4.6\% in All ACC and 6.1\% in New ACC.
In particular, SelEx+EAGC achieves state-of-the-art performance on all three datasets, with average improvements of 8.5\% in All ACC and 11.4\% in New ACC, demonstrating that EAGC substantially enhances the capability of SelEx.

\noindent\textbf{Results on Generic Datasets.} 
The results on generic datasets are presented on the right side of \cref{tab:sota}. 
Across the two generic benchmarks, incorporating EAGC yields average gains of 1.3\% in All ACC and 1.4\% in New ACC. 
The best results on CIFAR-100 and ImageNet-100 are achieved by SPTNet+EAGC and SelEx+EAGC, respectively. 
These results demonstrate that EAGC, as a plug-and-play module, effectively improves existing GCD baselines and leads to more accurate category discovery.

\noindent\textbf{Additional Results.}
Using DINOv2 as the backbone, EAGC consistently improves two baselines across three fine-grained datasets, with average gains of 6.2\% in All ACC and 8.3\% in New ACC. 
Results on the more challenging Herbarium19 dataset are provided in Appendix~B.

\begin{table}[!h]
\centering
\caption{Comparison with DINOv2 as the backbone.}
\setlength{\tabcolsep}{1.3pt}
\renewcommand{\arraystretch}{0.85}
\begin{tabular}{@{}lccccccccc@{}}
\toprule
\multirow{2}{*}{Method} & \multicolumn{3}{c}{CUB} & \multicolumn{3}{c}{Stanford Cars} & \multicolumn{3}{c}{Aircraft} \\
\cmidrule(lr){2-4}
\cmidrule(lr){5-7}
\cmidrule(lr){8-10}
 & All & Old & New & All & Old & New & All & Old & New \\
\midrule
SimGCD
& 71.5 & 78.1 & 68.3
& 71.5 & 81.9 & 66.6
& 63.9 & 69.9 & 60.9 \\

\cellcolor{my_blue}\quad + EAGC
& \cellcolor{my_blue}\textbf{77.9} & \cellcolor{my_blue}\textbf{79.9} & \cellcolor{my_blue}\textbf{76.9}
& \cellcolor{my_blue}\textbf{81.5} & \cellcolor{my_blue}\textbf{90.9} & \cellcolor{my_blue}\textbf{76.9}
& \cellcolor{my_blue}\textbf{74.6} & \cellcolor{my_blue}\textbf{74.9} & \cellcolor{my_blue}\textbf{74.4} \\

\midrule

SelEx
& 87.4 & \textbf{85.1} & 88.5
& 82.2 & \textbf{93.7} & 76.7
& 79.8 & \textbf{82.3} & 78.6 \\

\cellcolor{my_blue}\quad + EAGC
& \cellcolor{my_blue}\textbf{89.4} & \cellcolor{my_blue}82.3 & \cellcolor{my_blue}\textbf{92.9}
& \cellcolor{my_blue}\textbf{85.3} & \cellcolor{my_blue}92.7 & \cellcolor{my_blue}\textbf{81.7}
& \cellcolor{my_blue}\textbf{84.9} & \cellcolor{my_blue}81.8 & \cellcolor{my_blue}\textbf{86.5} \\
\midrule
Avg. $\Delta$
& \textcolor{ForestGreen}{+4.2} & \textcolor{red!50!white}{-0.5} & \textcolor{ForestGreen}{+6.5}
& \textcolor{ForestGreen}{+6.6} & \textcolor{ForestGreen}{+4.0} & \textcolor{ForestGreen}{+7.7}
& \textcolor{ForestGreen}{+7.9} & \textcolor{ForestGreen}{+2.3} & \textcolor{ForestGreen}{+10.7} \\
\bottomrule
\end{tabular}
\end{table}

\subsection{Ablation Study}
To evaluate the effectiveness of the AGA and EEP components in our EAGC framework, 
we perform ablation studies on both parametric and non-parametric baselines using the CUB and Stanford Cars datasets. 
The results are reported in \cref{tab:ablation_study}. 
\textbf{(1) AGA.} By anchoring the gradient updates of labeled samples, AGA effectively stabilizes the optimization of known categories. 
Incorporating AGA (variant \uppercase\expandafter{\romannumeral1}) yields average Old ACC improvements of 4.8\% for SimGCD and 3.0\% for SelEx. Interestingly, AGA can also improve the discovery of novel classes by providing more stable and transferable feature representations.
\textbf{(2) EEP.} By decoupling known and novel category subspaces at the gradient level, 
EEP (variant \uppercase\expandafter{\romannumeral2}) enhances the discovery accuracy of novel categories, 
achieving average New ACC gains of 4.6\% for SimGCD and 6.2\% for SelEx across the two datasets. In addition, the reduced subspace interference also benefits known-class recognition, yielding an average Old ACC gain of 2.7\%.
\textbf{(3) Full EAGC.} Combining AGA and EEP under the respective optimization objectives $\mathcal{L}_{\text{sup}}$ and $\mathcal{L}_{\text{unsup}}$ achieves the best overall results. 
Compared with variant \uppercase\expandafter{\romannumeral1}, EAGC improves All ACC by 1.6\% on average across the two datasets, 
and compared with variant \uppercase\expandafter{\romannumeral2}, it brings an additional 3.6\% gain in All ACC. These results demonstrate that AGA and EEP act in a complementary manner, jointly improving both known- and novel-class performance.
\begin{table}[!t]
\centering
\caption{
    Ablation study of the proposed components. 
    \textit{AGA} denotes Anchor-based Gradient Alignment, 
    and \textit{EEP} denotes Energy-aware Elastic Projection.
}
\label{tab:ablation_study}
\setlength{\tabcolsep}{3pt}
\renewcommand{\arraystretch}{0.9}
\begin{tabular}{@{}ccc|ccc|ccc@{}}
\toprule
\multirow{2}{*}{Method} 
& \multirow{2}{*}{\textit{AGA}} 
& \multirow{2}{*}{\textit{EEP}} 
& \multicolumn{3}{c|}{CUB} 
& \multicolumn{3}{c}{Stanford-Cars}  \\
\cmidrule(lr){4-6} 
\cmidrule(lr){7-9}
&  &  & All & Old & New & All & Old & New \\
\midrule
SimGCD    
& \textcolor{red}{\textbf{\sffamily x}} 
& \textcolor{red}{\textbf{\sffamily x}} 
& 60.3 & 65.6 & 57.7
& 53.8 & 71.9 & 45.0 \\

\midrule

\uppercase\expandafter{\romannumeral1}  
& \textcolor{mydarkgreen}{\checkmark} 
& \textcolor{red}{\textbf{\sffamily x}} 
& 65.6 & \textbf{71.1} & 62.9
& 62.2 & 75.9 & 55.6 \\

\uppercase\expandafter{\romannumeral2}  
& \textcolor{red}{\textbf{\sffamily x}} 
& \textcolor{mydarkgreen}{\checkmark} 
& 65.2  & \textbf{71.1}  & 62.2 
& 57.7 & 74.6 & 49.6 \\

+ EAGC  
& \textcolor{mydarkgreen}{\checkmark} 
& \textcolor{mydarkgreen}{\checkmark}
& \textbf{66.5} & 71.0 & \textbf{64.3}
& \textbf{62.9} & \textbf{76.0} & \textbf{56.6}\\

\midrule
SelEx    
& \textcolor{red}{\textbf{\sffamily x}} 
& \textcolor{red}{\textbf{\sffamily x}} 
& 73.6 & 75.3 & 72.8
& 58.5 & 75.6 & 50.3\\

\midrule

\uppercase\expandafter{\romannumeral1}  
& \textcolor{mydarkgreen}{\checkmark} 
& \textcolor{red}{\textbf{\sffamily x}} 
& 81.7 & 73.5 & 85.9 
& 62.6 & 83.4 & 52.6 \\

\uppercase\expandafter{\romannumeral2}  
& \textcolor{red}{\textbf{\sffamily x}} 
& \textcolor{mydarkgreen}{\checkmark} 
& 82.3 & 71.9 & 87.6 
& 58.8 & 81.5 & 47.9 \\

+ EAGC 
& \textcolor{mydarkgreen}{\checkmark} 
& \textcolor{mydarkgreen}{\checkmark}
& \textbf{83.2} & \textbf{73.9} & \textbf{87.9}
& \textbf{65.7} & \textbf{83.7} & \textbf{57.0}\\

\bottomrule
\end{tabular}
\end{table}


\subsection{Gradient Entanglement Analysis} 
We evaluate the effectiveness of EAGC in mitigating Gradient Entanglement, as summarized in \cref{tab:grad_entanglement}. 
The \textit{Gradient Deviation Coefficient} (\(\mathbb{GDC}\)) measures how unlabeled optimization distorts the gradient direction of labeled samples, 
whereas the \textit{Subspace Overlap Coefficient} (\(\mathbb{SOC}\)) quantifies the feature subspace overlap between known and novel categories. 
For fair comparison, both the baseline and our method adopt the same model configuration, and all metrics are computed as the average over the first 200 training steps. 
Across all benchmarks, integrating EAGC consistently lowers both \(\mathbb{GDC}\) and \(\mathbb{SOC}\), 
highlighting its effectiveness in decoupling the optimization dynamics of labeled and unlabeled data. 
For the parametric baseline SimGCD, EAGC achieves average reductions of 46.8\% in \(\mathbb{GDC}\) and 21.4\% in \(\mathbb{SOC}\), 
while for the non-parametric baseline SelEx, the reductions are 98.5\% and 18.1\%, respectively.

\begin{table}[!t]
\centering
\setlength{\tabcolsep}{1.1pt}
\renewcommand{\arraystretch}{1}
\caption{
Quantitative comparison of the Gradient Deviation Coefficient (\(\mathbb{GDC}\)) and Subspace Overlap Coefficient (\(\mathbb{SOC}\)).
}
\label{tab:grad_entanglement}
\begin{tabular}{@{}lccccccc@{}}
\toprule
\multirow{2}{*}{Method} 
& \multicolumn{2}{c}{CUB} 
& \multicolumn{2}{c}{Stanford Cars} 
& \multicolumn{2}{c}{Aircraft} \\
\cmidrule(lr){2-3}
\cmidrule(lr){4-5}
\cmidrule(lr){6-7}
 & \(\mathbb{GDC}\)$\downarrow$ & \(\mathbb{SOC}\)$\downarrow$ & \(\mathbb{GDC}\)$\downarrow$ & \(\mathbb{SOC}\)$\downarrow$ & \(\mathbb{GDC}\)$\downarrow$ & \(\mathbb{SOC}\)$\downarrow$ \\
\midrule

SimGCD~\cite{simgcd}
& 0.2669 & 0.6478 & 0.0909 & 0.8110 & 0.0935 & 0.8643 \\

\cellcolor{my_blue}\quad +EAGC
& \cellcolor{my_blue}\textbf{0.1525} & \cellcolor{my_blue}\textbf{0.4923} & \cellcolor{my_blue}\textbf{0.0500} & \cellcolor{my_blue}\textbf{0.6208} & \cellcolor{my_blue}\textbf{0.0375} & \cellcolor{my_blue}\textbf{0.7139} \\

\midrule

LegoGCD~\cite{legogcd}
& 0.2646 & 0.6468 & 0.0905 & 0.8112 & 0.0922 & 0.8642 \\

\cellcolor{my_blue}\quad +EAGC
& \cellcolor{my_blue}\textbf{0.1594} & \cellcolor{my_blue}\textbf{0.4844} & \cellcolor{my_blue}\textbf{0.0515} & \cellcolor{my_blue}\textbf{0.6135} & \cellcolor{my_blue}\textbf{0.0376} & \cellcolor{my_blue}\textbf{0.7112} \\

\midrule

SelEx~\cite{selex}
& 0.0005 & 0.5902 & 0.0007 & 0.6022 & 0.0313 & 0.6033 \\

\cellcolor{my_blue}\quad +EAGC
& \cellcolor{my_blue}\textbf{0.0003} & \cellcolor{my_blue}\textbf{0.4463} & \cellcolor{my_blue}\textbf{0.0001} & \cellcolor{my_blue}\textbf{0.4470} & \cellcolor{my_blue}\textbf{0.0001} & \cellcolor{my_blue}\textbf{0.5774} \\

\bottomrule
\end{tabular}
\end{table}

\subsection{Evaluation}
\noindent \textbf{Gradient Projection Strategy Evaluation.} 
Our Energy-aware Elastic Projection (EEP) adaptively assigns projection weights to unlabeled samples based on their energy proportions within the known-class subspace. 
We evaluate this design in \cref{tab:projection}, using the best-performing baseline SelEx as the base model.
The variant \textit{w/o Energy Adaptation} uniformly projects all unlabeled gradients away from the known-class subspace without considering sample-specific energy alignment. 
This uniform projection results in inferior performance, with an average drop of 4.2\% in All ACC. 
In particular, the accuracy on known classes decreases by 4.7\% on average, indicating that indiscriminate gradient projection hampers the learning of known samples within the unlabeled set.

\begin{table}[!h]
\centering
\setlength{\tabcolsep}{2pt}
\renewcommand{\arraystretch}{0.9}
\caption{
Evaluation of the gradient projection strategy based on SelEx. 
“w/o EEP” indicates a variant that uniformly projects all unlabeled gradients away from the known-class subspace without applying energy-aware weighting.
}
\label{tab:projection}
\begin{tabular}{@{}ccccccccccc@{}}
\toprule
\multirow{2}{*}{Method} 
& \multicolumn{3}{c}{CUB} 
& \multicolumn{3}{c}{Stanford Cars} 
& \multicolumn{3}{c}{Aircraft} \\
\cmidrule(lr){2-4}
\cmidrule(lr){5-7}
\cmidrule(lr){8-10}
 & All & Old & New & All & Old & New & All & Old & New \\
\midrule

$w/o$ EEP
& 80.9 & \textbf{74.5} & 84.2 
& 59.3 & 76.6 & 50.9 
& 61.7 & 57.9 & 63.6 \\

\cellcolor{my_blue}Ours
& \cellcolor{my_blue}\textbf{83.2} & \cellcolor{my_blue}73.9 & \cellcolor{my_blue}\textbf{87.9} 
& \cellcolor{my_blue}\textbf{65.7} & \cellcolor{my_blue}\textbf{83.7} & \cellcolor{my_blue}\textbf{57.0} 
& \cellcolor{my_blue}\textbf{65.7} & \cellcolor{my_blue}\textbf{65.6} & \cellcolor{my_blue}\textbf{65.7}\\

\bottomrule
\end{tabular}
\end{table}

\noindent \textbf{Known-class Representation Subspace Evaluation.} 
In our Energy-aware Elastic Projection, we employ a Conceptor-based soft subspace instead of a PCA-based hard subspace. 
While PCA captures only the top-$K$ principal components, it overlooks low-energy directions that may encode subtle discriminative cues between known and novel classes. 
We evaluate this design choice in \cref{tab:subspace}, where the number of principal components $K$ varies from 16 to 64. 
All PCA-based variants yield inferior overall results compared with our Conceptor-based formulation.
In particular, ours consistently improves performance on novel classes, surpassing the PCA-16 variant by an average of 1.2\%.

\begin{table}[!h]
\centering
\setlength{\tabcolsep}{6pt}
\renewcommand{\arraystretch}{0.9}
\caption{Comparison of subspace construction strategies. “PCA-K” refers to a hard subspace built from the top-$K$ principal components of labeled data.}
\label{tab:subspace}
\begin{tabular}{@{}ccccccc@{}}
\toprule
\multirow{2}{*}{Method} & \multicolumn{3}{c}{CUB} & \multicolumn{3}{c}{Stanford Cars} \\
\cmidrule(lr){2-4} 
\cmidrule(lr){5-7} 
 & All & Old & New & All & Old & New \\
\midrule
PCA-16 
& 82.7 & \textbf{74.5} & 86.8
& 64.9 & \textbf{84.0} & 55.7 \\

PCA-32 
& 80.9 & 73.5 & 84.7
& 64.5 & 84.1 & 55.0 \\

PCA-64
& 82.1 & 73.7 & 86.3
& 64.0 & 83.3 & 54.6 \\

\cellcolor{my_blue}Ours
& \cellcolor{my_blue}\textbf{83.2} & \cellcolor{my_blue}73.9 & \cellcolor{my_blue}\textbf{87.9} 
& \cellcolor{my_blue}\textbf{65.7} & \cellcolor{my_blue}83.7 & \cellcolor{my_blue}\textbf{57.0} \\
\bottomrule
\end{tabular}
\end{table}

\subsection{Hyperparameter Analysis}
\label{sec:hyper}

\begin{figure}[h!]
  \centering
  \includegraphics[width=\linewidth]{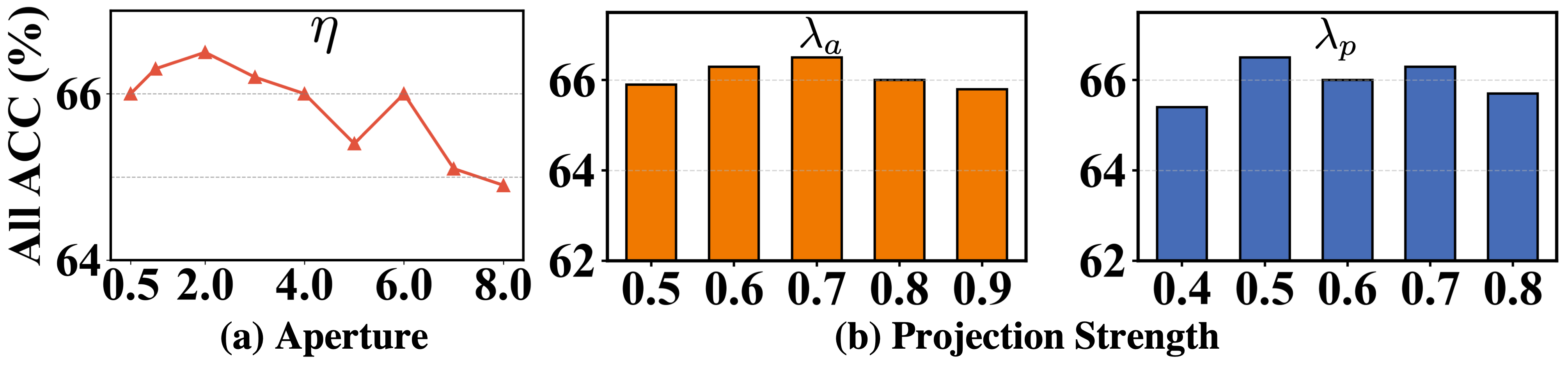}
 \caption{Effects of the aperture $\eta$ and projection-strength hyperparameters $\lambda_a$ and $\lambda_p$.}
    \label{fig:hyper1}
\end{figure}

\noindent \textbf{Aperture.} 
When constructing the known-class subspace, the aperture parameter controls the \textit{softness} of the subspace—larger aperture values correspond to softer subspaces that capture a broader range of directions. 
This is a shared hyperparameter across the entire EAGC framework. 
We determine the aperture based on experiments with the parametric baseline SimGCD on CUB and keep it fixed for all other datasets and baselines. 
As shown in \cref{fig:hyper1} (a), the performance remains stable across a wide range of aperture values (0.5–4.0). We set $\eta$ = 2.0 in all experiments.

\noindent \textbf{Projection Strength.} 
Our EAGC introduces two hyperparameters, $\lambda_a$ and $\lambda_p$, which control the projection strengths of AGA and EEP, respectively. 
To avoid over-tuning, we select them on CUB using the SimGCD baseline and keep them fixed for all baselines and datasets. 
The selection process is illustrated in \cref{fig:hyper1} (b) and (c). Overall, both $\lambda_a$ and $\lambda_p$ are relatively stable.
We set $\lambda_a$ = 0.7 and $\lambda_p$ = 0.5.

\section{Conclusion}

In this work, we identify and quantify a previously underexplored bottleneck in GCD, termed Gradient Entanglement, which fundamentally limits existing methods.
To tackle this issue, we propose the Energy-Aware Gradient Coordinator (EAGC), a plug-and-play gradient-level framework that stabilizes labeled gradients to preserve discriminative directions and adaptively decouples unlabeled gradients from the known-class subspace.
Extensive experiments across multiple benchmarks show that EAGC consistently improves diverse GCD baselines and achieves new state-of-the-art performance.
We expect EAGC to offer a new perspective for advancing GCD and to serve as a general strategy for future GCD methods.

\clearpage
\newpage
\section*{Acknowledgements}
This work was funded by the National Natural Science Foundation of China (No. 62572166, No. 62402157), and the Fundamental Research Funds for the Central Universities (No. JZ2025HGTB0219). This work was also supported by the EU Horizon projects ELIAS (No. 101120237) and ELLIOT (No. 101214398), as well as the FIS project GUIDANCE (No. FIS2023-03251). We further acknowledge CINECA and the ISCRA initiative for providing high-performance computing resources.

{
    \small
    \bibliographystyle{ieeenat_fullname}
    \bibliography{main}
}
\setcounter{page}{1}
\maketitlesupplementary
\renewcommand{\thesection}{\Alph{section}}

\setcounter{section}{0}
\setcounter{figure}{0}    
\setcounter{table}{0}   

\renewcommand{\thetable}{\Alph{table}}
\renewcommand{\thefigure}{\Alph{figure}}
\renewcommand{\thesection}{\Alph{section}}

\section*{Appendix Contents}
\begin{itemize}
  \item \hyperref[sec:supp_A]{\textbf{§A. Dataset Statistics}}  
  \item \hyperref[sec:supp_B]{\textbf{§B. Extended Experimental Results}}
    \begin{itemize}
      \item \hyperref[sec:supp_B1]{B.1 Computational overhead and scalability}
      \item \hyperref[sec:supp_B2]{B.2 Effectiveness on the Herbarium19 Dataset}
      \item \hyperref[sec:supp_B3]{B.3 Results with Unknown $K$}
      \item \hyperref[sec:supp_B4]{B.4 Exploration of Reference Models}
    \end{itemize}
  \item \hyperref[sec:supp_C]{\textbf{C. Theory}}
    \begin{itemize}
      \item \hyperref[sec:supp_C1]{C.1 Motivation from Proximal Regularization}
      \item \hyperref[sec:supp_C2]{C.2 Proximal Interpretation of AGA}
      \item \hyperref[sec:supp_C3]{C.3 Theoretical Properties}
      \item \hyperref[sec:supp_C4]{C.4 Empirical Exploration and Validation}
      \item \hyperref[sec:supp_C5]{C.5 Empirical Validation of Gradient Entanglement Hypotheses}
    \end{itemize}
  \item \hyperref[sec:supp_D]{\textbf{D. Broader Impact and Limitations Discussion}}
  \item \hyperref[sec:supp_E]{\textbf{E. Qualitative Analysis}}

\end{itemize}

\section{Dataset Statistics}
\label{sec:supp_A}
We follow the standard Generalized Category Discovery (GCD)~\cite{gcd} protocol to split the known/novel classes and determine the number of samples in the labeled and unlabeled subsets. We evaluate on four fine-grained datasets—CUB~\cite{cub}, Stanford Cars~\cite{scars}, Aircraft~\cite{aircraft}, and Herbarium19~\cite{herb19}—and two generic datasets, CIFAR-100~\cite{cifar} and ImageNet-100~\cite{imagenet}. For most datasets, 50\% of the classes are designated as known, with the exception of CIFAR-100, which adopts an 80\%/20\% known/novel split. Following the standard protocol, the labeled subset is constructed by sampling half of the images from the known classes. The unlabeled subset consists of all remaining images, which includes the other half of the known-class samples and all samples from the novel classes. Detailed statistics for all six datasets are provided in \cref{tab:dataset}.

\begin{table}[h]
\centering
\caption{Dataset statistics and splits.\label{tab:dataset}}
\setlength{\tabcolsep}{1pt}
\renewcommand{\arraystretch}{1}
\begin{tabular}{lrrrr}
\toprule
\multirow{2}{*}{Dataset} & \multicolumn{2}{c}{Labeled} & \multicolumn{2}{c}{Unlabeled} \\
\cmidrule(lr){2-3}\cmidrule(l){4-5}
& \#~Images & \#~Classes & \#~Images & \#~Classes \\
\midrule
CUB~\cite{cub}                     &  1,498 & 100  &  4,496 & 200 \\
Stanford Cars~\cite{scars}         &  2,000 &  98  &  6,144 & 196 \\
Aircraft~\cite{aircraft}         &  1,666 &  50  &  5,001 & 100 \\
CIFAR\mbox{-}100~\cite{cifar}      & 20,000 &  80  & 30,000 & 100 \\
ImageNet\mbox{-}100~\cite{imagenet}& 31,860 &  50  & 95,255 & 100 \\
Herbarium19~\cite{herb19}         &  8,869 &  341  &  25,356 & 683 \\
\bottomrule
\end{tabular}
\end{table}


\section{Extended Experimental Results}
\label{sec:supp_B}

\subsection{Computational overhead and scalability}
\label{sec:supp_B1}
In this section, we provide a detailed analysis of the computational overhead and scalability of the proposed Energy-Aware Gradient Coordinator (EAGC). First, it is important to note that EAGC is exclusively a training-time module. It incurs \textbf{zero computational cost during inference}. During training, the overhead of EAGC can be evaluated across three main aspects: memory consumption, matrix inversion efficiency, and time overhead. All empirical profiling is conducted and averaged over three independent runs on an NVIDIA A100 GPU, with results summarized in \cref{tab:overhead}.

\begin{table}[h]
\centering
\caption{Computational overhead analysis on an NVIDIA A100.}
\label{tab:overhead}
\setlength{\tabcolsep}{0.4pt}
\renewcommand{\arraystretch}{1}
\footnotesize
\begin{tabular}{c|l|c|c|c|c|c}
\toprule
\multirow{2}{*}{\textbf{Data}} & 
\multirow{2}{*}{\textbf{Method}} & 
\multirow{2}{*}{\textbf{All ACC}} & 
\textbf{Peak Mem} & 
\textbf{Inv. Cost} &
\multicolumn{2}{c}{\textbf{Time Costs (s)}} \\
& & & (MB) & (\textbf{ms/ep}) & Avg. Epoch & $\mathcal{E}_r(\cdot)$ Constr. \\
\midrule

\multirow{2}{*}{\textbf{CUB}}
 & SimGCD & 60.3 & 4289.4 & - & 55.9 & - \\
 & + \textbf{\textit{EAGC}} & 66.5 & 4927.2 & 2.9 & 63.8 & 467.9 \\ 

\midrule
\multirow{2}{*}{\textbf{IN-100}}
 & SimGCD & 83.0 & 4288.1 & - & 947.7 & - \\
 & + \textbf{\textit{EAGC}} & 83.5 & 4921.3 & 3.5 & 1262.2 & 6899.1 \\ 

\bottomrule
\end{tabular}
\end{table}

\noindent\textbf{Memory Overhead.} EAGC requires a slight increase in GPU memory (e.g., from 4289 MB to 4927 MB on the CUB dataset). This increase is primarily due to loading the frozen reference model $\mathcal{E}_r(\cdot)$ to extract reference features. Since $\mathcal{E}_r(\cdot)$ is kept strictly frozen, we do not need to construct or store its computational graph for gradients, ensuring that the memory overhead remains highly manageable.

\noindent\textbf{Scalability of Matrix Inversion.} The matrix inversion step in the elastic projection scales efficiently with the feature dimension. Profiling on ImageNet-100 reveals that the inversion cost is 3.5 ms/epoch for a feature dimension of $d=768$. Increasing the dimension to $d=1024$ yields an inversion cost of 7.3 ms/epoch. This demonstrates that matrix inversion adds only a small overhead to the training process, confirming the scalability of EAGC to higher-dimensional representations.

\noindent\textbf{Training Time versus Performance Gain.} The time overhead introduced by EAGC consists of two components: a one-time construction cost for the reference model $\mathcal{E}_r(\cdot)$ prior to the main training phase, and the per-epoch gradient coordination cost. As shown in \cref{tab:overhead}, on the CUB dataset, the one-time construction takes 467.9 seconds, and the per-epoch overhead is approximately 7.9 seconds (a 14.1\% increase relative to the SimGCD baseline). Overall, the method effectively translates this 14.1\% increase in training time into a 6.2\% absolute gain in All ACC on the CUB dataset.

\subsection{Effectiveness on the Herbarium19 Dataset}
\label{sec:supp_B2}
To further evaluate the generalizability and robustness of our proposed EAGC on large-scale, long-tailed distributions, we conduct extended experiments on the challenging Herbarium19 dataset. As shown in \cref{tab:herb19_results}, integrating EAGC consistently improves the performance of both the parametric baseline (SimGCD) and the non-parametric baseline (SelEx). Specifically, EAGC yields an average absolute improvement of 3.4\% in All ACC and 4.9\% in New ACC across the two baselines. Notably, when applied to SimGCD, EAGC significantly boosts the New ACC by 6.0\%. This demonstrates the effectiveness and robustness of our gradient coordination mechanism.

\begin{table}[!h]
\centering
\setlength{\tabcolsep}{5pt}
\renewcommand{\arraystretch}{0.9}
\caption{Extended results on the long-tailed Herbarium19 dataset.}
\label{tab:herb19_results}
\begin{tabular}{@{}ccccccc@{}}
\toprule
\multirow{2}{*}{Method} & \multicolumn{3}{c}{SimGCD} & \multicolumn{3}{c}{SelEx} \\
\cmidrule(lr){2-4} 
\cmidrule(lr){5-7} 
 & All & Old & New & All & Old & New \\
\midrule
Baseline
& 44.0 & 58.0 & 36.4
& 39.6 & 54.9 & 31.3 \\

\cellcolor{my_blue}\quad+ EAGC
& \cellcolor{my_blue}\textbf{48.4}  & \cellcolor{my_blue}\textbf{59.8} & \cellcolor{my_blue}\textbf{42.4}
& \cellcolor{my_blue}\textbf{42.0}  & \cellcolor{my_blue}\textbf{55.0} & \cellcolor{my_blue}\textbf{35.0} \\

\bottomrule
\end{tabular}
\end{table}

\subsection{Results with Unknown $K$}
\label{sec:supp_B3}
Our EAGC is a plug-and-play module applied during GCD model training and can also be used when the number of novel classes is unknown. Using the class-number estimation method from GCD~\cite{gcd}, we evaluate the effect of integrating EAGC into the baselines on CUB, Stanford Cars, and ImageNet-100. The results are shown in \cref{tab:wok}. Specifically, on CUB (estimated 231 vs.\ ground-truth 200 classes), Stanford Cars (estimated 230 vs.\ 196), and ImageNet-100 (estimated 108 vs.\ 100), our EAGC consistently improves the baselines—boosting SimGCD by an average of 6.6\% and SelEx by 6.7\% in All ACC. Notably, on the two fine-grained datasets, even under large estimation errors in the number of classes, EAGC significantly enhances the novel-class discovery performance, yielding an average improvement of 10.9\% in New ACC across both baselines.

\begin{table}[!h]
\centering
\small
\setlength{\tabcolsep}{1pt}
\renewcommand{\arraystretch}{0.9}
\caption{Comparison without Number of Categories $K$\label{tab:wok}}
\begin{tabular}{@{}ccccccccccc@{}}
\toprule
\multirow{2}{*}{Method} 
& \multicolumn{3}{c}{CUB} 
& \multicolumn{3}{c}{Stanford Cars} 
& \multicolumn{3}{c}{ImageNet-100} \\
\cmidrule(lr){2-4}
\cmidrule(lr){5-7}
\cmidrule(lr){8-10}
 & All & Old & New & All & Old & New & All & Old & New \\
\midrule

SimGCD~\cite{simgcd}
& 61.5 & 66.4 & 59.1
& 49.1 & 65.1 & 41.3
& 81.7 & 91.2 & 76.8 \\

\cellcolor{my_blue}\quad +EAGC
&\cellcolor{my_blue}66.5  &\cellcolor{my_blue}69.4  &\cellcolor{my_blue}65.0  
&\cellcolor{my_blue}63.3  &\cellcolor{my_blue}74.8  &\cellcolor{my_blue}\textbf{57.8}  
&\cellcolor{my_blue}82.2  &\cellcolor{my_blue}94.1  &\cellcolor{my_blue}76.2 \\

\midrule

SelEx~\cite{selex}
& 72.0 & 72.3 & 71.9
& 58.7 & 75.3 & 50.8
& 85.4 & 94.0 & 81.0 \\

\cellcolor{my_blue}\quad +EAGC
&\cellcolor{my_blue}\textbf{83.1}  &\cellcolor{my_blue}\textbf{75.4}  &\cellcolor{my_blue}\textbf{86.9}  
&\cellcolor{my_blue}\textbf{64.2}  &\cellcolor{my_blue}\textbf{79.1}  &\cellcolor{my_blue}56.9  
&\cellcolor{my_blue}\textbf{88.9}  &\cellcolor{my_blue}\textbf{95.5}  &\cellcolor{my_blue}\textbf{85.5} \\

\midrule

\textbf{Avg. $\triangle$ }
&\textcolor{ForestGreen}{+8.1}&\textcolor{ForestGreen}{+3.1}&\textcolor{ForestGreen}{+10.5}&\textcolor{ForestGreen}{+9.9}&\textcolor{ForestGreen}{+6.8}&\textcolor{ForestGreen}{+11.3}&\textcolor{ForestGreen}{+2.0}&\textcolor{ForestGreen}{+2.2}&\textcolor{ForestGreen}{+2.0}\\

\bottomrule
\end{tabular}
\end{table}

\subsection{Exploration of the Reference Model}
\label{sec:supp_B4}
Our AGA relies on a reference model trained in a supervised manner on the labeled subset to perform gradient alignment. Here, we conduct an extensive analysis of the reference model.

\noindent \textbf{1) Training-related hyperparameters.}  
In EAGC, we use a batch size of 32 for fine-grained datasets and 128 for generic datasets, and we train for 30 epochs on all datasets.  
\emph{i) Batch size.}  
GCD training typically uses mixed labeled and unlabeled data within a mini-batch with a batch size of 128. However, for supervised training on the labeled subset, we find that using a smaller batch size yields better results for fine-grained datasets. As shown in \cref{tab:pbs}, using a batch size of 32 reduces $\mathcal{L}_{\text{cls}}$ by an average of 0.18 across the three datasets and improves All ACC by an average of 3.3\% compared to a batch size of 128.  
\emph{ii) Training epoch.}  
We evaluate training for 20, 30, and 100 epochs, with results shown in \cref{tab:pepoch}. Overall, too few epochs may lead to underfitting on known classes, while too many epochs can cause overfitting. We therefore choose a balanced setting and fix the training schedule to 30 epochs unless otherwise specified.

\vspace{-0.2cm}
\begin{table}[!h]
\footnotesize
\centering
\setlength{\tabcolsep}{1pt}
\renewcommand{\arraystretch}{0.7}
\caption{Effect of batch size for training the reference model.}
\label{tab:pbs}
\begin{tabular}{@{}c|cccc|cccc|cccc@{}}
\toprule
Batch
& \multicolumn{4}{c|}{CUB} 
& \multicolumn{4}{c|}{Stanford Cars} 
& \multicolumn{4}{c}{Aircraft} \\
\cmidrule(lr){2-5}
\cmidrule(lr){6-9}
\cmidrule(lr){10-13}
 Size&$\mathcal{L}_{\text{cls}}$ & All & Old & New &$\mathcal{L}_{\text{cls}}$ & All & Old & New &$\mathcal{L}_{\text{cls}}$ & All & Old & New \\
\midrule

\cellcolor{my_blue}32
&\cellcolor{my_blue}0.021 & \cellcolor{my_blue}66.5 & \cellcolor{my_blue}71.0 & \cellcolor{my_blue}64.3
&\cellcolor{my_blue}0.034 & \cellcolor{my_blue}\textbf{62.9} & \cellcolor{my_blue}\textbf{76.0} & \cellcolor{my_blue}\textbf{56.6}
&\cellcolor{my_blue}0.042 & \cellcolor{my_blue}\textbf{57.7} & \cellcolor{my_blue}\textbf{60.4} & \cellcolor{my_blue}\textbf{56.3}\\

128
& 0.035 & \textbf{68.6} & \textbf{71.3} & \textbf{67.3}
& 0.052 & 58.2 & 71.8 & 51.7
& 0.554 & 50.3 & 58.3 & 46.3\\

\bottomrule
\end{tabular}
\end{table}

\vspace{-0.2cm}
\begin{table}[!h]
\footnotesize
\centering
\setlength{\tabcolsep}{1pt}
\renewcommand{\arraystretch}{0.7}
\caption{Effect of the number of training epochs for the reference model.}
\label{tab:pepoch}
\begin{tabular}{@{}c|cccc|cccc|cccc@{}}
\toprule
Training
& \multicolumn{4}{c|}{CUB} 
& \multicolumn{4}{c|}{Stanford Cars} 
& \multicolumn{4}{c}{Aircraft} \\
\cmidrule(lr){2-5}
\cmidrule(lr){6-9}
\cmidrule(lr){10-13}
 Epoch&$\mathcal{L}_{\text{cls}}$ & All & Old & New &$\mathcal{L}_{\text{cls}}$ & All & Old & New &$\mathcal{L}_{\text{cls}}$ & All & Old & New \\
\midrule

20
& 0.046 & \textbf{67.8} & \textbf{71.8} & \textbf{65.9}
& 0.130 & 55.1 & 71.8 & 47.0
& 0.232 & 55.6 & \textbf{60.9} & 52.9 \\

\cellcolor{my_blue}30
&\cellcolor{my_blue}0.021 & \cellcolor{my_blue}66.5 & \cellcolor{my_blue}71.0 & \cellcolor{my_blue}64.3
&\cellcolor{my_blue}0.034 & \cellcolor{my_blue}\textbf{62.9} & \cellcolor{my_blue}\textbf{76.0} & \cellcolor{my_blue}\textbf{56.6}
&\cellcolor{my_blue}0.042 & \cellcolor{my_blue}57.7 & \cellcolor{my_blue}60.4 & \cellcolor{my_blue}56.3\\

100
& 0.014 & 65.7 & 67.5 & 64.8
& 0.018 & 54.9 & 69.0 & 48.1
& 0.011 & \textbf{57.9} & 58.3 & \textbf{57.7} \\

\bottomrule
\end{tabular}
\end{table}

\vspace{-0.2cm}
\noindent \textbf{2) Exploring trainable parameters.}  
In EAGC, we \textbf{follow the baselines' configuration for the reference model's trainable parameters}, e.g., setting the number of unfrozen blocks to 1 for SimGCD and 2 for SelEx. We further explore the effect of trainable parameters using SimGCD as the baseline. As shown in \cref{tab:pblock}, using more trainable parameters helps the model better learn known classes, resulting in lower $\mathcal{L}_{\text{cls}}$.

\vspace{-0.2cm}
\begin{table}[!h]
\small
\centering
\setlength{\tabcolsep}{4pt}
\renewcommand{\arraystretch}{0.7}
\caption{Effect of the number of unfrozen ViT blocks in the reference model.}
\label{tab:pblock}

\begin{tabular}{@{}c|cccc|cccc@{}}
\toprule

Unfrozen 
& \multicolumn{4}{c|}{CUB} 
& \multicolumn{4}{c}{Stanford Cars} \\
\cmidrule(lr){2-5}
\cmidrule(lr){6-9}
Blocks & $\mathcal{L}_{\text{cls}}$ & All & Old & New 
      & $\mathcal{L}_{\text{cls}}$ & All & Old & New \\
\midrule

\cellcolor{my_blue}1
&\cellcolor{my_blue}0.021 & \cellcolor{my_blue}66.5 & \cellcolor{my_blue}71.0 & \cellcolor{my_blue}64.3
&\cellcolor{my_blue}0.034 & \cellcolor{my_blue}62.9 & \cellcolor{my_blue}\textbf{76.0} & \cellcolor{my_blue}56.6 \\

2
& 0.015 & 65.5 & \textbf{73.3} & 61.5
& 0.028 & 56.9 & 74.9 & 48.2 \\

3
& 0.013 & 66.2 & 71.2 & 63.7
& 0.025 & 60.6 & 72.1 & 55.1 \\

4
& 0.012 & \textbf{67.3} & 69.6 & \textbf{66.1}
& 0.016 & \textbf{66.4} & 74.7 & \textbf{62.4} \\

5
& 0.012 & 66.6 & 69.7 & 65.1
& 0.019 & 57.7 & 71.4 & 51.1 \\

\bottomrule
\end{tabular}
\end{table}

\section{Theory}
\label{sec:supp_C}

\subsection{Motivation from Proximal Regularization}
\label{sec:supp_C1}
In GCD, the joint objective $\mathcal{L}_{\text{GCD}} = \alpha \mathcal{L}_{\text{sup}} + \beta \mathcal{L}_{\text{unsup}}$ combines a relatively reliable supervised signal from labeled data with a much noisier unsupervised signal driven by self-supervision or pseudo-labels. 
This intrinsic asymmetry means that the overall gradient can easily deviate from the desirable supervised direction, especially in the early training stage where $\mathcal{L}_{\text{unsup}}$ is highly unstable. 
\textit{Proximal regularization}~\cite{nocedal2006numerical,yun2021adaptive} is a classical tool for stabilizing optimization under noisy or conflicting gradients, by penalizing deviations from a trusted reference point and thereby damping harmful updates. 
Viewed through this lens, it is natural to interpret our Anchor-based Gradient Alignment (AGA) as introducing a proximal force that keeps the optimization of labeled samples close to a supervised anchor, while still allowing the model to benefit from unlabeled objectives.  

\subsection{Proximal Interpretation of AGA}
\label{sec:supp_C2}
Given the reference feature $\hat{z}^l$ obtained from the reference model $\mathcal{E}_r(\cdot)$, 
AGA introduces the following alignment term for labeled samples:
\begin{equation}
\nabla_{z^l} g_{\text{align}} = \lambda_a (z^l - \hat{z}^l).
\end{equation}
This term corresponds exactly to the gradient of a proximal regularizer that constrains the labeled representation $z^l$ to remain within a neighborhood of the reliable supervised optimum $\hat{z}^l$.  
Under this view, the effective objective for labeled data becomes:
\begin{equation}
\mathcal{L}_{\text{sup}}'
= \mathcal{L}_{\text{sup}}
+ \frac{\lambda_a}{2} \| z^l - \hat{z}^l \|_2^2,
\label{eq:sup_mse}
\end{equation}
which is a classical proximal step widely used in numerical optimization and robust learning.  
Therefore, AGA can be rigorously interpreted as a \textit{feature-space proximal point update}, 
where the reference feature $\hat{z}^l$ serves as the proximal center and $\lambda_a$ governs the radius of the trust region.  
This viewpoint clarifies that AGA does not override unsupervised learning; instead, it stabilizes supervised learning by regularizing labeled features toward a reliable optimum, thereby suppressing gradient distortion induced by $\mathcal{L}_{\text{unsup}}$.  
In our EAGC framework, AGA serves as a principled proximal regularizer for labeled-sample gradients under a unified gradient-optimization perspective.

\subsection{Theoretical Properties}
\label{sec:supp_C3}
We next show that viewing AGA as a proximal regularizer implies two beneficial properties under mild local assumptions.

\noindent\textit{Lemma 1 (Gradient variance reduction).}
Consider the labeled feature $z^l$ in a local neighborhood of the supervised optimum $\hat z^l$, and assume that the supervised loss \textit{admits a second–order approximation}
\begin{equation}
\mathcal{L}_{\text{sup}}(z^l) \approx \mathcal{L}_{\text{sup}}(\hat z^l)
+ \tfrac{1}{2}(z^l-\hat z^l)^\top H (z^l-\hat z^l),
\end{equation}
where $H \succeq 0$ denotes the local Hessian. 
We model the interference from the unsupervised objective as a 
stochastic disturbance $\xi$ acting on the shared backbone’s gradient 
pathway. Following standard analyses of noise-driven gradient 
perturbations, we assume
\begin{equation}
\mathbb{E}[\xi]=0,\quad \mathrm{Cov}[\xi]=\Sigma.
\end{equation}
Then, under a step size that ensures stable local updates around the supervised optimum and the quadratic approximation above, the steady–state covariance of the labeled–feature gradient with AGA is no larger (in the PSD sense) than that without AGA:
\begin{equation}
\mathrm{Cov}\!\big[\nabla_{z^l}\mathcal{L}^{\prime}_{\text{sup}}(z^l)\big]
\;\preceq\;
\mathrm{Cov}\!\big[\nabla_{z^l}\mathcal{L}_{\text{sup}}(z^l)\big],
\end{equation}
where $\mathcal{L}^{\prime}_{\text{sup}}(z^l)=\mathcal{L}_{\text{sup}}(z^l)+\frac{\lambda_a}{2}\|z^l-\hat z^l\|_2^2$ denotes the proximal-regularized objective.

\noindent \textit{Proof.}
Under the quadratic approximation, the supervised gradient is
\begin{equation}
\nabla_{z^l}\mathcal{L}_{\text{sup}}(z^l) = H (z^l-\hat z^l).
\end{equation}
A single feature–level gradient step \emph{without} AGA (ignoring higher–order terms) can be written as
\begin{equation}
z^l_{t+1} = z^l_t - \eta\big(H(z^l_t-\hat z^l) + \xi_t\big),
\end{equation}
where $t\in\{0,1,2,\dots\}$ denotes the iteration index of gradient descent, $\eta>0$ is the learning rate and $\xi_t$ is the stochastic perturbation from the unlabeled objective. 
Defining $\delta_t = z^l_t - \hat z^l$, this becomes a linear stochastic system:
\begin{equation}
\delta_{t+1} = (I - \eta H)\delta_t - \eta \xi_t.
\end{equation}
Assuming the Markov chain is stable, the stationary covariance $\Sigma_\delta = \mathrm{Cov}[\delta_t]$ satisfies (up to $\mathcal{O}(\eta^2)$ terms):
\begin{equation}
H \,\Sigma_\delta + \Sigma_\delta H \;\approx\; \eta\,\Sigma.
\end{equation}
Consequently, the covariance of the supervised gradient is:
\begin{equation}
\mathrm{Cov}\!\big[\nabla_{z^l}\mathcal{L}_{\text{sup}}(z^l)\big]
= \mathrm{Cov}[H\delta_t]
\approx H \Sigma_\delta H.
\end{equation}
With AGA, the effective objective becomes
\begin{equation}
\mathcal{L}^{\prime}_{\text{sup}}(z^l)
= \mathcal{L}_{\text{sup}}(z^l)
+ \tfrac{\lambda_a}{2}\|z^l-\hat z^l\|_2^2,
\end{equation}
whose gradient is
\begin{equation}
\nabla_{z^l}\mathcal{L}^{\prime}_{\text{sup}}(z^l)
= (H+\lambda_a I)(z^l-\hat z^l).
\end{equation}
The corresponding feature dynamics (again in the local quadratic regime) are
\begin{equation}
\delta_{t+1} = (I - \eta (H+\lambda_a I))\delta_t - \eta \xi_t.
\end{equation}
By the same argument, the stationary covariance $\Sigma_\delta^{\prime}$ now satisfies
\begin{equation}
(H+\lambda_a I)\Sigma_\delta^{\prime} 
+ \Sigma_\delta^{\prime}(H+\lambda_a I)
\;\approx\; \eta\,\Sigma,
\end{equation}
and thus
\begin{equation}
\mathrm{Cov}\!\big[\nabla_{z^l}\mathcal{L}^{\prime}_{\text{sup}}(z^l)\big]
\approx (H+\lambda_a I)\,\Sigma_\delta^{\prime}\,(H+\lambda_a I).
\end{equation}

\noindent Solving the Lyapunov equations in both cases yields the well–known      ``ridge'' effect:
\begin{equation}
\Sigma_\delta^{\prime} 
\;\preceq\;
\Sigma_\delta,\quad
(H+\lambda_a I)^{-1}\Sigma(H+\lambda_a I)^{-1}
\;\preceq\;
H^{-1}\Sigma H^{-1},
\end{equation}
whenever $\lambda_a>0$ and $H\succeq 0$.  
Equivalently,
\begin{equation}
\mathrm{Cov}\!\big[\nabla_{z^l}\mathcal{L}^{\prime}_{\text{sup}}(z^l)\big]
\;\preceq\;
\mathrm{Cov}\!\big[\nabla_{z^l}\mathcal{L}_{\text{sup}}(z^l)\big].
\end{equation}
\hfill$\square$

\subsection{Empirical Exploration and Validation}
\label{sec:supp_C4}
Building on SelEx~\cite{selex}, which achieves the best performance, we design two sets of experiments to examine the role of proximal regularization:
(i) we directly adopt $\mathcal{L}_{\text{sup}}'$ in \cref{eq:sup_mse} as an explicit optimization objective for labeled data and disable the AGA module within EAGC;
(ii) we replace AGA with relation-based distillation losses.

\noindent\textit{(i) Loss-based proximal regularization.}
As shown in \cref{tab:proximal_loss_vs_aga}, when we replace AGA with $\mathcal{L}_{\text{sup}}'$ as the optimization objective for labeled data, the overall performance is almost identical to that of the original AGA: compared with the SelEx baseline, we obtain an average improvement of about 7.6\% in All ACC on the two datasets.
This result validates our design from two perspectives:
first, AGA essentially implements, at the gradient level, the proximal regularizer induced by $\mathcal{L}_{\text{sup}}'$;
second, proximal regularization with respect to a reference model can substantially boost overall GCD performance.

\begin{table}[!t]
\centering
\setlength{\tabcolsep}{5pt}
\renewcommand{\arraystretch}{0.75}
\caption{Evaluation on loss-based proximal regularization (replacing gradient-level AGA with the explicit loss $\mathcal{L}_{\text{sup}}'$)}
\label{tab:proximal_loss_vs_aga}

\begin{tabular}{@{}lcccccc@{}}
\toprule
\multirow{2}{*}{Method} & \multicolumn{3}{c}{CUB} & \multicolumn{3}{c}{Stanford Cars} \\
\cmidrule(lr){2-4} 
\cmidrule(lr){5-7} 
 & All & Old & New & All & Old & New \\
\midrule

SelEx~\cite{selex}
& 73.6 & 75.3 & 72.8
& 58.5 & 75.6 & 50.3 \\

\cellcolor{my_blue}\quad +EAGC
&\cellcolor{my_blue} \textbf{83.2} &\cellcolor{my_blue} 73.9 &\cellcolor{my_blue} \textbf{87.9} 
&\cellcolor{my_blue} \textbf{65.7} &\cellcolor{my_blue} 83.7 &\cellcolor{my_blue} \textbf{57.0} \\
\midrule
\quad +EAGC$_{\text{loss}}$
& 82.1 & \textbf{74.5} & 85.9 
& 65.1 & \textbf{83.9} & 56.1 \\
\bottomrule
\end{tabular}
\end{table}

\begin{figure}[h]
  \centering
  \vspace{-0.3cm}
  \includegraphics[width=0.9\linewidth]{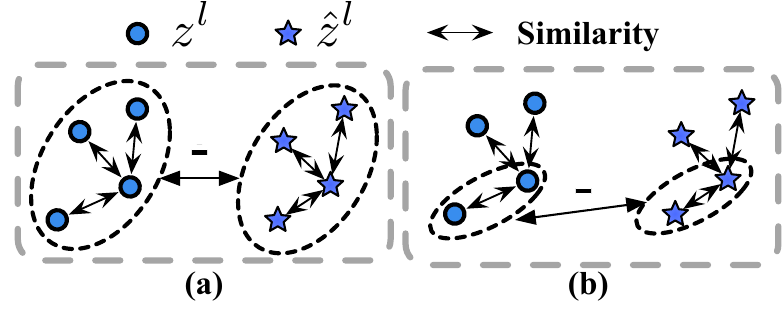}

\caption{Comparison of two relation-based distillation variants: (a) Gram-matrix distillation; (b) sample-level relation distillation.}
\vspace{-0.2cm}
\label{fig:appendix_distill}
\end{figure}

\begin{table}[!h]
\centering
\setlength{\tabcolsep}{3pt}
\renewcommand{\arraystretch}{0.7}
\caption{Comparison of relation-based distillation variants and gradient-level AGA.}
\label{tab:relational_distill_variants}
\begin{tabular}{@{}lcccccc@{}}
\toprule
\multirow{2}{*}{Method} & \multicolumn{3}{c}{CUB} & \multicolumn{3}{c}{Stanford Cars} \\
\cmidrule(lr){2-4} 
\cmidrule(lr){5-7} 
 & All & Old & New & All & Old & New \\
\midrule

SelEx~\cite{selex}
& 73.6 & 75.3 & 72.8
& 58.5 & 75.6 & 50.3 \\

\cellcolor{my_blue}\quad +EAGC
&\cellcolor{my_blue} \textbf{83.2} &\cellcolor{my_blue} 73.9 &\cellcolor{my_blue} \textbf{87.9} 
&\cellcolor{my_blue} \textbf{65.7} &\cellcolor{my_blue} \textbf{83.7} &\cellcolor{my_blue} \textbf{57.0} \\
\midrule
\quad +RelDistill$_{\text{Gram}}$
& 80.4 & 68.5 & 86.3
& 59.8 & 79.7 & 50.1 \\

\quad +RelDistill$_{\text{sample}}$
& 82.9 & \textbf{79.5} & 84.6
& 60.0 & 81.9 & 49.5 \\

\bottomrule
\end{tabular}
\end{table}

\noindent \textit{(ii) Relation-based distillation in place of proximal regularization.}
This experiment has two aims: (i) to verify whether a reference model trained solely on labeled data can provide effective structural guidance for known classes; and (ii) to assess to what extent different forms of constraints can substitute for AGA.
As illustrated in~\cref{fig:appendix_distill}, we construct two relation-based distillation variants on top of the reference model:
(a) Gram-matrix distillation, which aligns the global relations among labeled samples (denoted as \texttt{+RelDistill$_{\text{Gram}}$});
and (b) sample-level relation distillation based on neighbor distributions (denoted as \texttt{+RelDistill$_{\text{sample}}$}).
The results are summarized in \cref{tab:relational_distill_variants}.
We observe that Gram-matrix distillation can enforce that the overall pairwise relations among labeled samples remain consistent with those of the reference model, but it cannot prevent global drift in the feature space; as a result, it still trails our AGA by an average of 4.4\% in All ACC across the two datasets.
In contrast, the sample-level relation distillation more effectively constrains the labeled samples and better preserves the model’s discriminative ability on known classes: across the two datasets, its All ACC is on average 3.0\% lower than AGA, while its Old ACC is 1.9\% higher.
Overall, both the gradient-level proximal regularization implemented by AGA and the relation-based distillation derived from the reference model can substantially strengthen the original SelEx baseline, albeit to different extents.

\subsection{Empirical Validation of Gradient Entanglement Hypotheses}
\label{sec:supp_C5}
To provide deeper insights into the optimization dynamics of Generalized Category Discovery (GCD) and empirically validate our core hypotheses, we conduct a detailed gradient analysis using the SimGCD baseline on the CUB dataset. Specifically, we aim to examine two key assumptions: first, that novel-class samples can induce gradient deviation; and second, that known-class gradients may dominate the optimization process, contributing to representation collapse.

\begin{figure}[h]
\centering
\includegraphics[width=\linewidth]{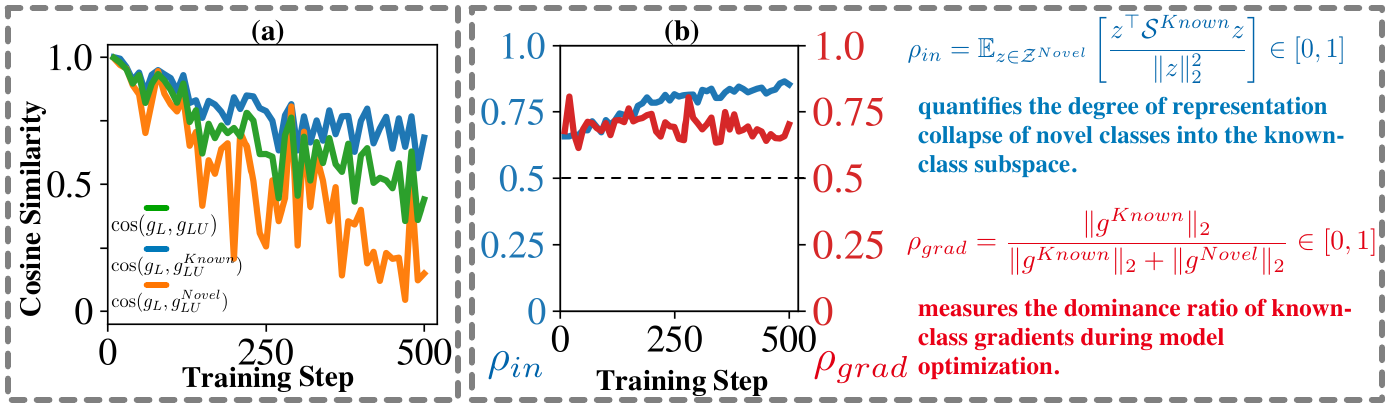}
\caption{Empirical validation of gradient entanglement. (a) Cosine similarity between the reference supervised gradient ($g_L$) and various mixed gradients. (b) The evolution of gradient dominance ($\rho_{\text{grad}}$) and representation collapse coefficient ($\rho_{\text{in}}$) during the joint training process.}
\label{fig:empirical_validation}
\end{figure}

\noindent\textbf{Impact of Novel Classes on Gradient Deviation.} We first analyze the gradient directions at the same parameter checkpoint under four distinct data settings: labeled data only ($g_L$), labeled combined with known-class unlabeled data ($g_{LU}^{Known}$), labeled combined with novel-class unlabeled data ($g_{LU}^{Novel}$), and the standard GCD setting using all data ($g_{LU}$). As illustrated in \cref{fig:empirical_validation} (a), the gradient $g_{LU}^{Known}$ maintains a high cosine similarity with the reference supervised gradient $g_L$. In contrast, the introduction of novel-class unlabeled samples (yielding $g_{LU}^{Novel}$ and $g_{LU}$) noticeably reduces this similarity, resulting in a clear angular deviation. This observation empirically supports our hypothesis that novel-class samples can directly distort the optimization direction of the supervised objective.

\noindent\textbf{Gradient Dominance and Representation Collapse.} Furthermore, we hypothesize that the optimization process is largely dominated by known-class gradients, which inadvertently pull novel-class representations into the known-class subspace. To validate this, we track two metrics throughout the training process: gradient dominance ($\rho_{\text{grad}}$), defined as the proportion of the total gradient norm contributed by known classes, and the representation collapse coefficient ($\rho_{\text{in}}$), which quantifies the extent to which novel features are projected into the known-class subspace. As shown in \cref{fig:empirical_validation} (b), $\rho_{\text{grad}}$ consistently remains above 0.5, confirming the dominance of known-class gradients. Concurrently, $\rho_{\text{in}}$ steadily increases as training progresses. This suggests that dominant known-class gradients progressively attract novel-class representations, thereby increasing subspace overlap and reducing the separability of novel categories.

\section{Broader Impact and Limitations Discussion}
\label{sec:supp_D}
\noindent \textbf{Broader Impact.}
This work improves category discovery in mixed known--novel settings from an optimization perspective, which may enhance the reliability of open-world visual systems when encountering unseen categories. In real-world applications, such capability is particularly relevant for unknown object discovery and dynamic environment understanding, where models are required to distinguish between known and previously unseen objects \cite{liang2023unknown,xue2024indoor}. In addition, more stable category discovery can facilitate the organization of large-scale unlabeled data, potentially reducing manual annotation efforts in applications such as industrial anomaly detection and novel pattern discovery \cite{huang2025anomalyncd}. Beyond the GCD setting, the proposed optimization perspective on mitigating interference between known and novel representations may also provide insights for broader open-world visual understanding tasks, including open-vocabulary and domain-generalized segmentation \cite{zhao2026open,zheng2026open,li2025maris,li2025fgaseg,li2026exploring}.

\noindent \textbf{Limitations.}
This method still has several limitations. First, EAGC relies to some extent on the quality of known-class representations, including the reference model introduced by AGA and the known-class subspace constructed by EEP. If these representations are not sufficiently accurate or stable, the effectiveness of gradient coordination may be affected. Second, although EAGC can be integrated into existing frameworks in a plug-and-play manner, it still introduces additional computational overhead, which may become a burden in large-scale training or resource-constrained scenarios. Finally, the current evaluation is mainly conducted on standard GCD benchmarks, and its performance in more complex open-world settings, such as larger category spaces, significant domain shifts, or dynamically changing environments, remains to be further studied.

\noindent \textbf{Future Work.}
There are several promising directions for future work. First, while the current method is evaluated on standard GCD benchmarks, it would be valuable to study its applicability in more complex open-world settings, such as scenarios with significant domain shifts, noisy pseudo-labels, or more dynamic and non-stationary environments \cite{zhao2025secov2,shen8437180ai,lei2026multi}. Second, as EAGC is largely decoupled from specific model architectures, it can be combined with stronger representation learning and pre-training frameworks to further improve the quality of known-class structures and enhance gradient coordination \cite{ren2023masked,ren2026masked}. Finally, it is also of interest to extend the proposed method to broader tasks, including multimodal understanding \cite{li2025lion,li2025cogvla}, 3D point cloud perception \cite{qu2024conditional,qu2025end}, and robotics-related applications \cite{yan2025pandas,ruan2024q}.

\section{Qualitative Analysis}
\label{sec:supp_E}

\begin{figure*}[!h]
  \centering
  \includegraphics[width=\linewidth]{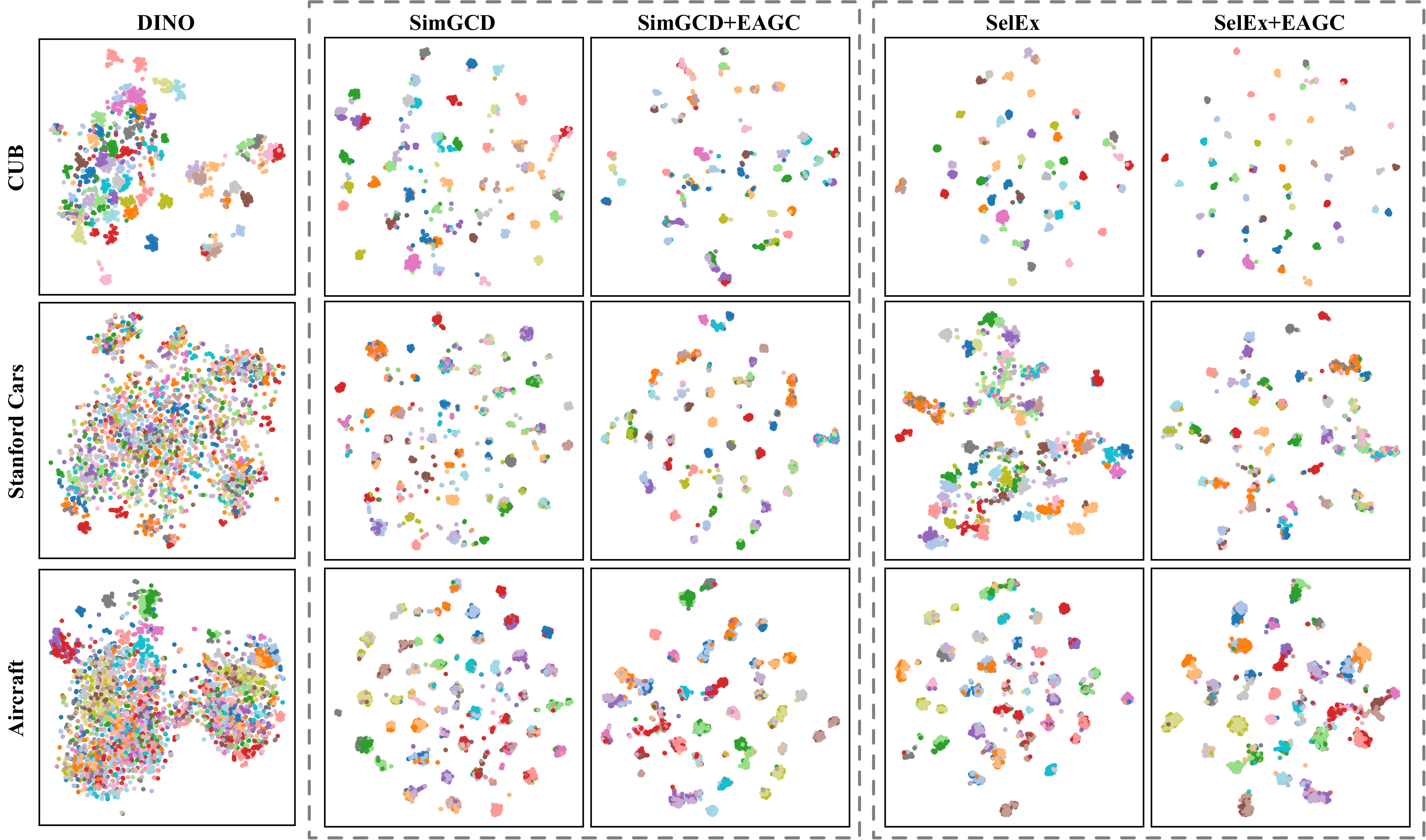}

\caption{Qualitative comparison using t-SNE visualizations.}
\label{fig:appendix_tsne}

\end{figure*}

\begin{figure*}[!h]
  \centering
  \includegraphics[width=\linewidth]{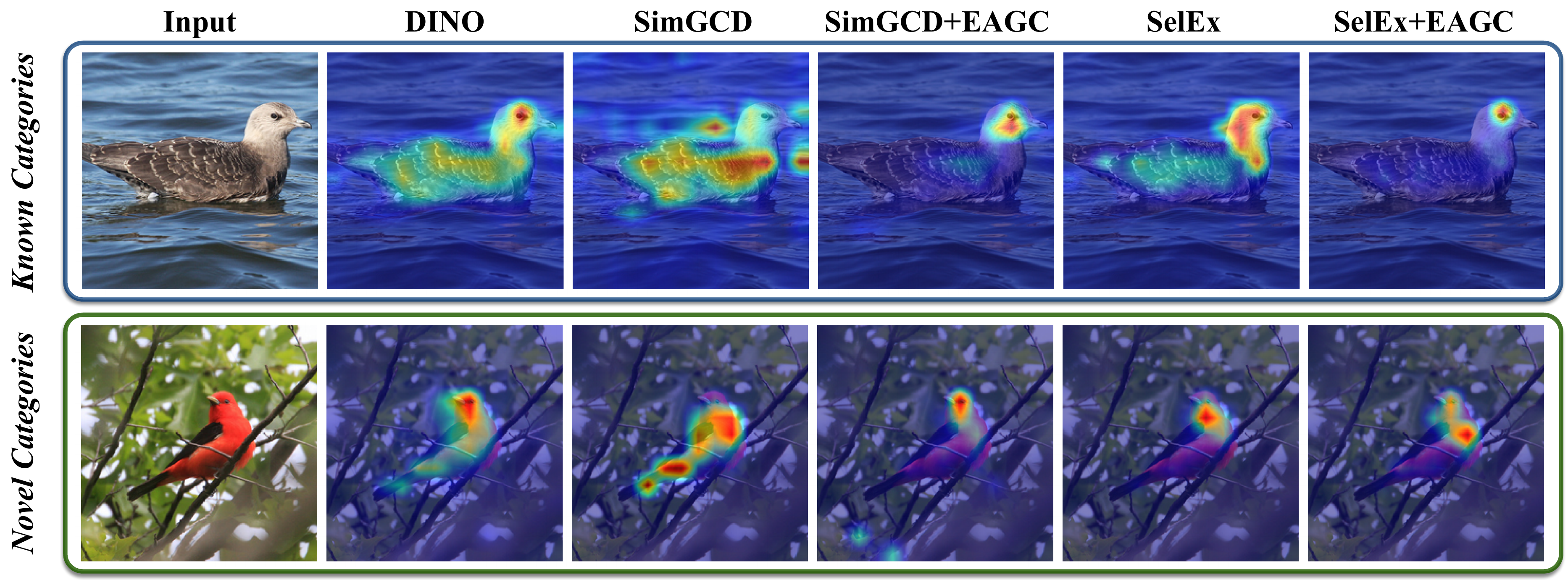}

\caption{Visualization of attention maps of baselines and EAGC-integrated baselines on CUB.}
\label{fig:appendix_attn_cub1}

\end{figure*}

\begin{figure*}[!h]
  \centering
  \includegraphics[width=0.9\linewidth]{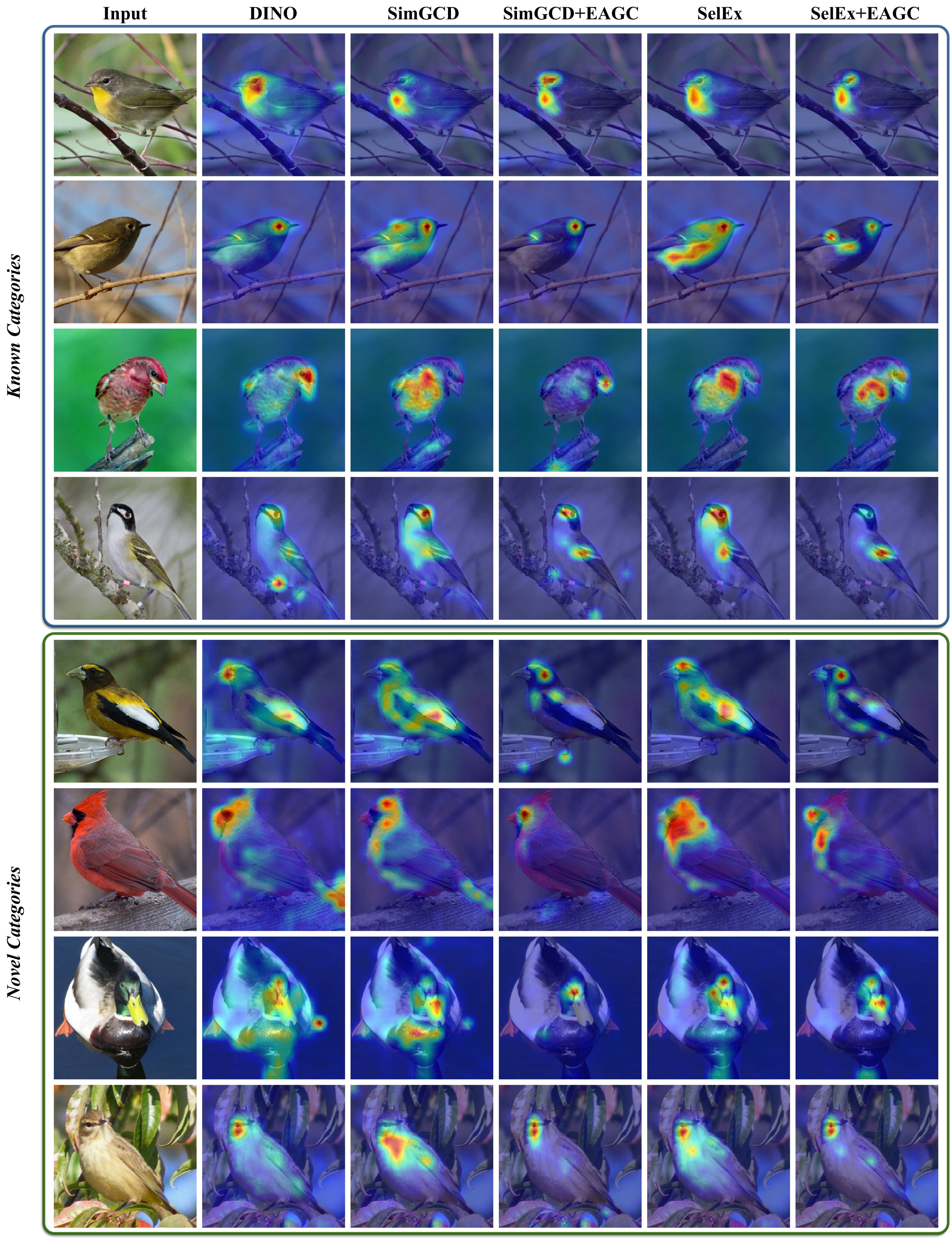}

\caption{Additional Visualization of attention maps of baselines and EAGC-integrated baselines on CUB.}
\label{fig:appendix_attn_cub2}

\end{figure*}

\begin{figure*}[!h]
  \centering
  \includegraphics[width=0.9\linewidth]{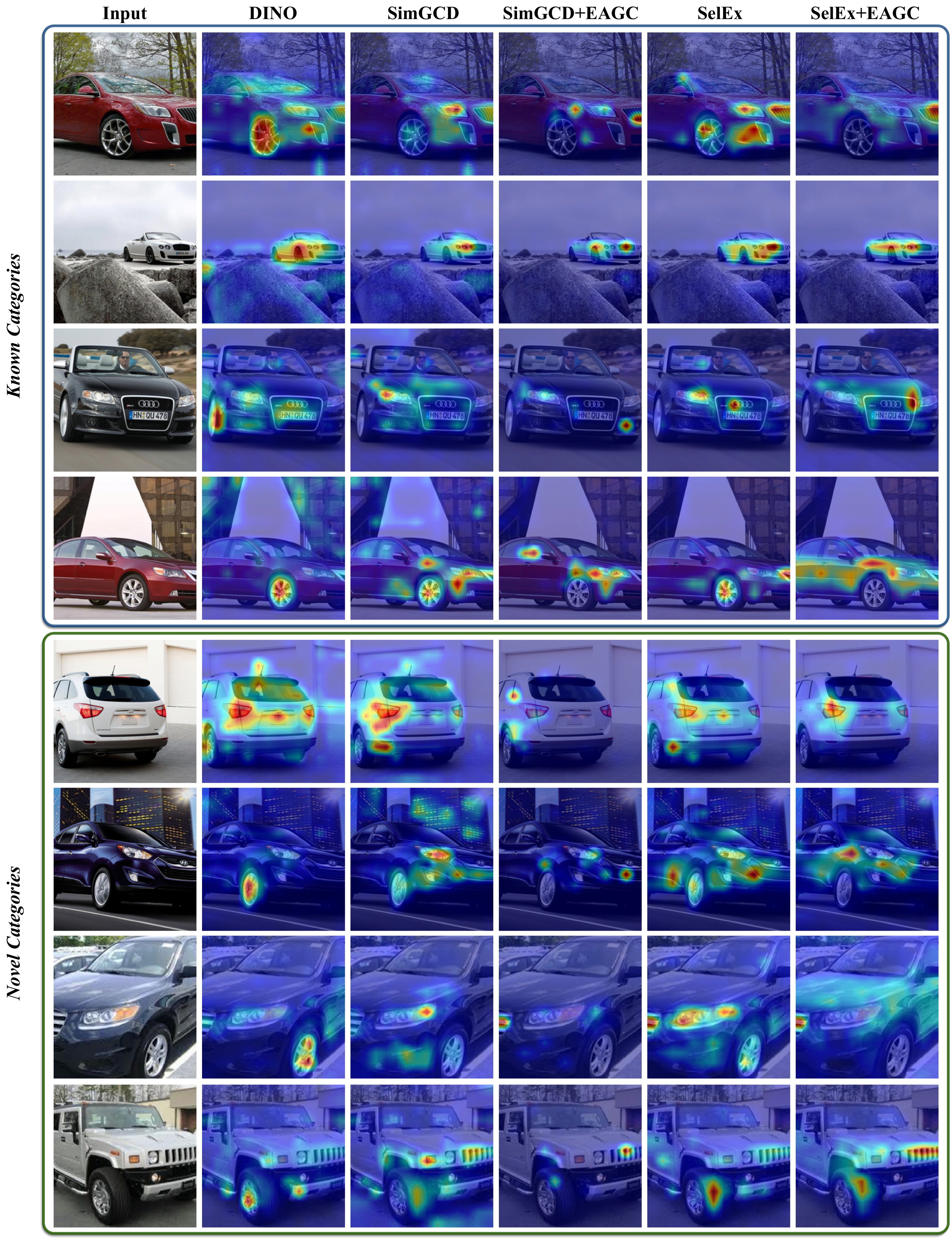}

\caption{Visualization of attention maps of baselines and EAGC-integrated baselines on Stanford Cars.}
\label{fig:appendix_attn_scars}

\end{figure*}

\begin{figure*}[!h]
  \centering
  \includegraphics[width=0.9\linewidth]{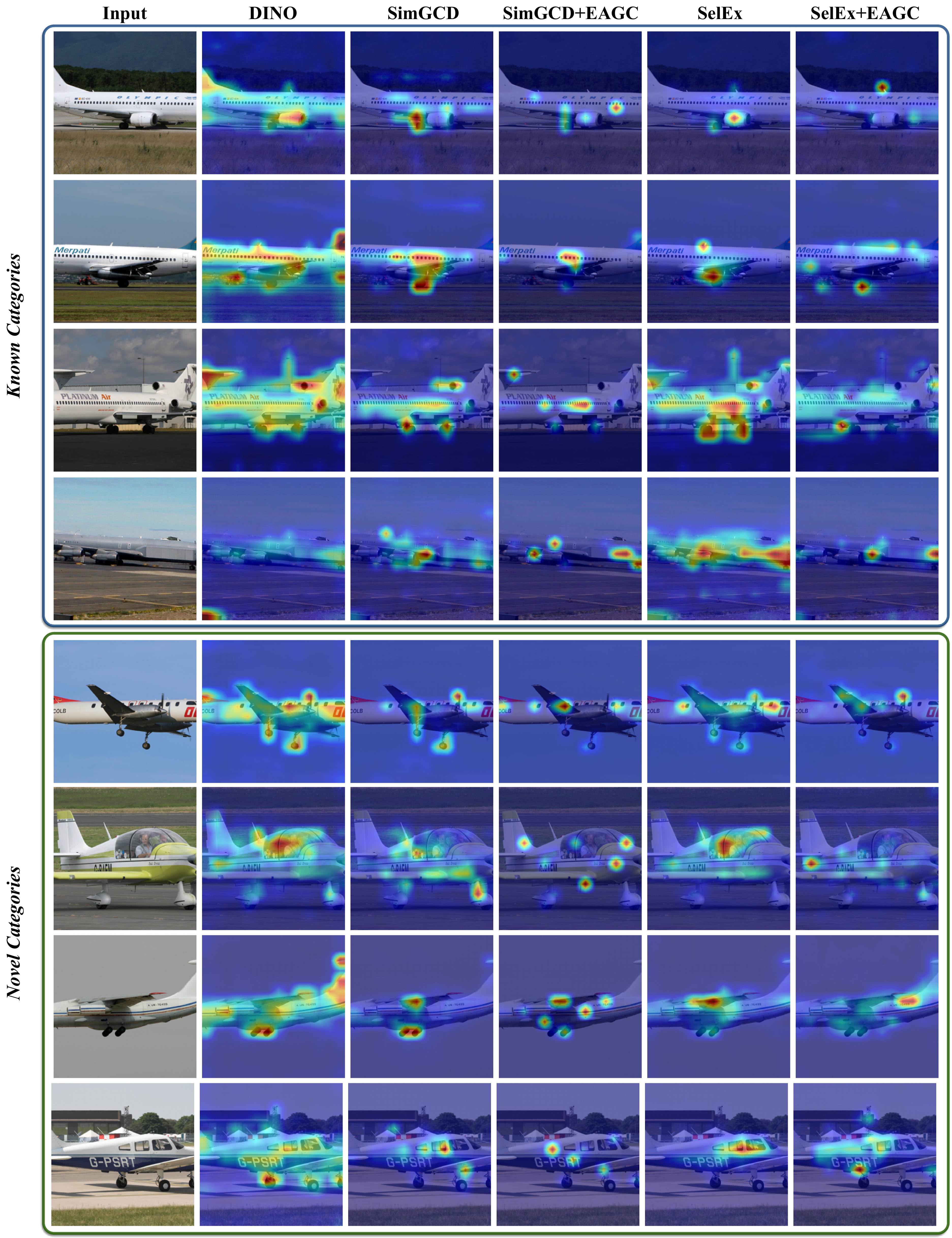}

\caption{Visualization of attention maps of baselines and EAGC-integrated baselines on Aircraft.}
\label{fig:appendix_attn_aircraft}

\end{figure*}

We present t-SNE visualizations of feature distributions across the CUB, Stanford Cars, and Aircraft datasets in \cref{fig:appendix_tsne}, comparing DINO, two baselines (SimGCD and SelEx), and these baselines integrated with our EAGC. To ensure visual clarity, we display only the last 50 classes (all belonging to novel categories) based on class indices. As illustrated, while the baselines demonstrate improved clustering capabilities over the raw DINO features, the incorporation of EAGC yields sharper boundaries and clearer inter-cluster separation. Furthermore, we also visualize the attention maps for DINO, baselines, and EAGC-integrated baselines in \cref{fig:appendix_attn_cub1,fig:appendix_attn_cub2,fig:appendix_attn_scars,fig:appendix_attn_aircraft}. Benefiting from stabilized gradient updates, our EAGC-integrated baselines demonstrate \textbf{more concentrated and precise attention}. The attention focuses on critical fine-grained details, such as bird eyes and beaks, car logos, and aircraft portholes.

\clearpage


\end{document}